\documentclass[journal]{IEEEtran}

\usepackage[english]{babel}

\usepackage{ifpdf}

\usepackage{cite} 
\usepackage{url}
\usepackage{hyperref}
\usepackage{tcolorbox}

\usepackage{color}
\usepackage{pgf, tikz, pgfplots}
\usetikzlibrary{shapes, arrows, automata, plotmarks}
\usetikzlibrary{calc,hobby,decorations}

\usepackage{tikz-cd}

\usepackage[cmex10]{amsmath}
\usepackage{amsfonts, amssymb, amsthm}
\usepackage{mathrsfs}

\usepackage{algorithm,algpseudocode}
	\algnewcommand{\LeftComment}[1]{\Statex \(\triangleright\) #1}

\newcommand{\makeboxlabel}[1]{#1\hfill}

\usepackage{enumerate}
\usepackage{multirow}
\usepackage{rotating}
\usepackage{subcaption}
\usepackage{needspace}

\input{./latex_resources/mySymbol.sty}
\input{./latex_resources/pennColors.sty}

\definecolor{my_cp_col1}{RGB}{253, 231, 37}
\definecolor{my_cp_col2}{RGB}{180, 222,44}
\definecolor{my_cp_col3}{RGB}{94, 201, 98}
\definecolor{my_cp_col4}{RGB}{33, 145, 140}
\definecolor{my_cp_col5}{RGB}{59, 82, 139}
\definecolor{my_cp_col6}{RGB}{68, 1, 84}

\definecolor{my_cp2_col1}{RGB}{255,255,217}
\definecolor{my_cp2_col2}{RGB}{237,248,177}
\definecolor{my_cp2_col3}{RGB}{199,233,180}
\definecolor{my_cp2_col4}{RGB}{127,205,187}
\definecolor{my_cp2_col5}{RGB}{65,182,196}
\definecolor{my_cp2_col6}{RGB}{29,145,192}
\definecolor{my_cp2_col7}{RGB}{34,94,168}
\definecolor{my_cp2_col8}{RGB}{37,52,148}
\definecolor{my_cp2_col9}{RGB}{8,29,88}

\newtheorem{example}{\hspace{0pt}\bf Example}

\newtheorem{theorem}{\hspace{0pt}\bf Theorem}
\newtheorem{corollary}{\hspace{0pt}\bf Corollary}

\newtheorem{remark}{\hspace{0pt}\bf Remark}

\newtheorem{definition}{\hspace{0pt}\bf Definition}

\usepackage{amsmath,amsfonts,graphicx, tikz,bm}

\usepackage{algpseudocode, algorithm,xcolor,amsthm}

\usepackage{ifthen}
\newboolean{showcomments}
\setboolean{showcomments}{true}
\usepackage{todonotes}

\usepackage{lipsum}

\definecolor{green}{RGB}{107,142,35}

\newcommand{\juan}[1]{  \ifthenelse{\boolean{showcomments}}
{\todo[inline,color=pink]{Juan: #1}}{}}

\newcommand{\alejandropm}[1]{  \ifthenelse{\boolean{showcomments}}
{\todo[inline,color=cyan]{AlejandroPM: #1}}{}}

\newcommand{\alejandror}[1]{  \ifthenelse{\boolean{showcomments}}
{\todo[inline,color=orange]{AlejandroR: #1}}{}}

\begin{document}

\title{RKHS Representation of Algebraic Convolutional Filters with Integral Operators}

\author{Alejandro Parada-Mayorga$^\spadesuit$, Alejandro Ribeiro$^{\small\triangle}$, and Juan Bazerque$^\diamondsuit$
\thanks{$\spadesuit$ Dept. of Electrical Eng., Univ.of Colorado (Denver). USA e-mail: \{alejandro.paradamayorga\}@ucdenver.edu. $\diamondsuit$ Dept. of Electrical and Comput Eng., Univ.of Pittsburgh. USA email: juanbazerque@pitt.edu. $\triangle$ Dept. of Electrical and Systems Eng., Univ.of Pennsylvania.}
}

\markboth{Signal Processing}
{Shell \MakeLowercase{\textit{et. al.}}: Bare Demo of IEEEtran.cls for Journals}

\maketitle




\begin{abstract}
Integral operators play a central role in signal processing, underpinning classical convolution, and filtering on continuous network models such as graphons. While these operators are traditionally analyzed through spectral decompositions, their connection to reproducing kernel Hilbert spaces (RKHS) has not been systematically explored within the algebraic signal processing framework. In this paper, we develop a comprehensive theory showing that the range of integral operators naturally induces RKHS convolutional signal models whose reproducing kernels are determined by a box product of the operator symbols. We characterize the algebraic and spectral properties of these induced RKHS and show that polynomial filtering with integral operators corresponds to iterated box products, giving rise to a unital kernel algebra. This perspective yields pointwise RKHS representations of filters via the reproducing property, providing an alternative to operator-based implementations. Our results establish precise connections between eigendecompositions and RKHS representations in graphon signal processing, extend naturally to directed graphons, and enable novel spatial--spectral localization results. Furthermore, we show that when the spectral domain is a subset of the original domain of the signals, optimal filters for regularized learning problems admit finite-dimensional RKHS representations, providing a principled foundation for learnable filters in integral-operator-based neural architectures. 
\end{abstract}


\IEEEpeerreviewmaketitle



\section{Introduction}

Convolutional signal processing has emerged as a foundational framework for analyzing and processing structured data across a wide range of domains, from classical time-series analysis to modern graph-based learning systems. The success of convolutional architectures—particularly in deep learning—has motivated the development of rigorous mathematical frameworks capable of unifying and extending filtering operations beyond traditional Euclidean domains. Two theoretical perspectives have proven especially influential in this effort: reproducing kernel Hilbert spaces (RKHS) and algebraic signal processing (ASP). Recent work~\cite{rkhs_conv} established that any RKHS naturally induces algebraic convolutional models, demonstrating how the representation property of RKHS—where functions admit expansions as linear combinations of kernel functions—can be leveraged to define convolution operations via products on kernel centers. When the underlying domain carries a monoid or group structure, these constructions give rise to RKHS algebras that support convolutional filtering on a broad class of spaces, including groups, graphons, and Euclidean domains.

While this framework provides powerful tools for constructing convolutions from RKHS, a complementary and equally fundamental question concerns the \emph{inverse direction}: under what conditions do classical filtering operations, implemented through integral operators, naturally induce RKHS structures? Integral operators are ubiquitous in signal processing. They define classical convolutions on $\mathbb{R}$, model diffusion processes on graphs through adjacency-based kernels, and characterize filtering on network limits such as graphons. A fundamental, yet underexplored, observation is that the \emph{range} of such integral operators acting on $L_{2}$ spaces naturally carries the structure of a reproducing kernel Hilbert space. Although this fact is well known in functional analysis, its implications for signal processing—and in particular for algebraic signal processing—have not been systematically developed.

The connection between integral operators and RKHS arises through the so called box product operation. Specifically, if $\boldsymbol{T}_S$ is an integral operator with symbol $S(u,v)$, acting according to $\boldsymbol{T}_{S}f=\int_{\ccalX}S(u,v)f(v)dv$, then its range induces an RKHS on $\ccalX$ with reproducing kernel
\[
K(u,v) = (S \square S^{\ast})(u,v)
        = \int S(u,z)\overline{S}(z,v)\,dz.
\]
Consequently, signals obtained by filtering through integral operators inherit an RKHS structure whose kernel is explicitly determined by the operator symbol. In contrast to our prior work~\cite{rkhs_conv}, which focused on constructing convolutional algebras from existing RKHS by exploiting algebraic structure on the domain, the present paper investigates how integral-operator-based filtering \emph{creates} RKHS structures and how these induced spaces can be exploited for filter representation, spectral analysis, and learning.

This perspective is particularly relevant in graphon signal processing (Gphon-SP), where the graphon shift operator $\boldsymbol{T}_{W}$ is itself an integral operator and polynomial filters correspond to compositions of such operators. Classical Gphon-SP studies these filters primarily through the eigendecomposition of $\boldsymbol{T}_W$. By contrast, the RKHS viewpoint developed here reveals that each polynomial diffusion produces signals lying in a specific RKHS with kernel $K=W\square W$ (for $p(\boldsymbol{T}_W)$), or more generally $K=W^{\square 2n}$ for $p(\boldsymbol{T}_W^{2n})$. This interpretation establishes a precise relationship between graphon eigenspaces and RKHS representations, and it extends naturally to directed graphons (digraphons), where symmetry is lost but the integral structure persists.

In this paper, we develop a comprehensive theory of RKHS structures induced by integral operators within the algebraic signal processing framework. This work complements~\cite{rkhs_conv}, which demonstrated how convolutional algebras arise from RKHS when the domain admits a monoid or group operation. Whereas that work emphasized algebraic structure on the domain, the present paper focuses on algebraic structure intrinsic to the integral operators themselves. Our analysis shows that integral operators induce a distinct algebraic framework—based on the box product—that supports polynomial filtering, spectral analysis, and learning directly in RKHS.

Our central contributions lie in characterizing the box product algebra induced by integral operators and in showing how this structure yields novel filter representations. We show here that iterated box products naturally generate a unital algebra in which polynomial operations correspond to polynomial filtering. We establish that filters themselves admit RKHS representations: a polynomial filter can be represented and implemented via point-wise inner products using the reproducing property of the underlying RKHS. This yields both computational advantages and conceptual insights distinct from operator-based filtering.

The main contributions of this work are summarized as follows:


\smallskip
\begin{list}
      {}
      {\setlength{\labelwidth}{22pt}
       \setlength{\labelsep}{0pt}
       \setlength{\itemsep}{0pt}
       \setlength{\leftmargin}{22pt}
       \setlength{\rightmargin}{0pt}
       \setlength{\itemindent}{0pt} 
       \let\makelabel=\makeboxlabel
       }

%
%
\item[{\bf (C1)}]\textit{Integral operators and induced RKHS:} We establish that integral operators with appropriate symbols induce reproducing kernel Hilbert spaces through their range spaces. For classical convolutions on $\mathbb{R}$, we show that bandlimited signal spaces arise naturally as RKHS (Section~II-A). In the context of graphon signal processing, we prove that the range of the graphon shift operator $\boldsymbol{T}_W$ determines an RKHS with kernel $K = W \square W$, and we characterize the precise relationship between graphon eigendecompositions and RKHS representations (Theorems~\ref{thm_K_from_W_gphonsp}--\ref{thm_W_gphon_from_K}). These results are further extended to digraphons, showing that polynomial diffusions admit RKHS-based spectral interpretations even in the absence of symmetry (Theorem~\ref{thm_diff_digraphon}).
\item[{\bf (C2)}]\textit{Box product algebra for integral operators:}
We introduce and analyze the box product operation on reproducing kernels, proving that it induces a unital algebra in which polynomial filtering corresponds to iterated box products (Theorem~\ref{thm_alg_RKHS_box_prod_1}). We derive spectral representations showing that polynomials in $K$ -- with the box product as underlying product -- admit eigenfunction expansions with coefficients given by polynomials of the associated eigenvalues (Theorem~\ref{thm_box_prod_spect}). This algebraic structure is intrinsic to integral operator–based models and is fundamentally different from the domain-induced convolution algebras developed in~\cite{rkhs_conv}.
\item[{\bf (C3)}] \textit{Point-wise filter representations in RKHS:}
We develop a homomorphism that realizes polynomial diffusions through point-wise inner products by exploiting the reproducing property (Theorem~\ref{thm_hom_box_prod}). We prove that this representation is equivalent to classical integral-operator implementations of filtering (Theorem~\ref{thm_alg_rkhs_vs_alg_classic}), yielding a novel computational paradigm in which filtering can be performed entirely within RKHS. Moreover, we show that filtered signals decompose as sums of RKHS spaces induced by iterated box products (Corollary~\ref{cor_diff_as_sum_Hkn}).
\item[{\bf (C4)}] \textit{Spatial--spectral localization for integral-operator filters:}
We establish fundamental tradeoffs between frequency bandlimitation and RKHS-finite duration (Corollary~\ref{cor_rkhs_uncertainty}), showing that the support size of an expansion on kernel evaluations imposes intrinsic limits on the achievable bandwidth. Using Corollary~\ref{cor_spec_vs_ku}, we characterize how coefficients in RKHS expansions determine spectral behavior through discrete inner products with eigenfunctions, providing explicit design principles for filters with prescribed frequency responses.
\item[{\bf (C5)}] \textit{Learnable filters in RKHS:}
We prove that optimal filters minimizing regularized error functionals on operator spectra admit finite-dimensional representations as expansions in kernel functions centered at eigenvalues (Corollary~\ref{cor_filt_representer}), when the spectrum of the kernel is a subset of the domain of the signals. This result provides a rigorous theoretical foundation for learning convolutional filters in integral-operator-based neural architectures, including graphon models.
\end{list}
\smallskip


\textbf{Organization.} The remainder of this paper is organized as follows. Section~\ref{sec_conv_int_opt_rkhs} establishes how integral operators induce reproducing kernel Hilbert space structures, with applications to classical convolution on $\mathbb{R}$, graphon signal processing, and digraphon models. Section~\ref{sec_alg_conv_filt_int_opt_rkhs} develops the algebraic signal processing theory of integral-operator-based filtering in RKHS, introducing the box product algebra and point-wise filter representations. Section~\ref{sec_rkhs_filt_rep_asp_filt_desgn} addresses RKHS-based filter design, including spatial--spectral localization results and the theory of learnable filters via representer theorems. Finally, Section~\ref{sec_discussion} concludes the paper and discusses future research directions.




\section{Convolutions with Integral Operators in RKHS}
\label{sec_conv_int_opt_rkhs}

A reproducing kernel Hilbert space (RKHS) is a Hilbert space of functions where the evaluation functional is bounded~\cite{paulsen2016introduction,wainwright2019high,ghojogh2021reproducing}. Therefore, if $\ccalH$ is an RKHS, we have $\ccalH\subset\ccalF(\ccalX,\mbC)$ where $\ccalF(\ccalX,\mbC)$ is the vector space of functions from an arbitrary domain $\ccalX$ to the complex numbers $\mbC$, and if $E_{u}:\ccalH\to\mbC$ is the evaluation functional $E_{v}(f)=f(v)$, it follows that $\vert E_{v}(f)\vert\leq C_{v}\Vert f\Vert_{\ccalH}$ for every $v\in\ccalX$ and any $f\in \ccalH$ with $C_{v}>0$~\cite{paulsen2016introduction,wainwright2019high,ghojogh2021reproducing}. Taking into account the Riez representation theorem~\cite{paulsen2016introduction}, it is possible to show that for any $v\in\ccalX$ there exists a $k_{v}(x)\in\ccalH$ such that $f(v)=\langle f, k_v \rangle_{\ccalH}$ for all $f\in\ccalH$. The function $k_{v}$ allows the definition of a positive semidefinite kernel $K: \ccalX\times\ccalX\to\mbC$ that encapsulates some properties of the RKHS. Such a kernel is obtained as $K(u,v)=k_{v}(u)$, and using $K(u,v)$ one can write any function $f$ in the RKHS $\ccalH$ as
\begin{equation}\label{eqn_representation_property}
   f(x) = \sum_{v} \alpha_{v} k_{v}(x)
        = \sum_{v} \alpha_v K(x, v) 
        ,
\end{equation}
where $\alpha_{v}\in\mbC$. This is the so-called representation property. Additionally, the inner product in $\ccalH$, endows $K$ with the so called reproducing property given by
\begin{equation}\label{eqn_reproducing_property}
   \big\langle k_v(x), k_u(x) \big\rangle_{\ccalH}
      = \big\langle K(x , v),  K(x, u)  \big\rangle_{\ccalH}
      =  K(u, v) 
      .
\end{equation}
A fundamental insight when combining~\eqref{eqn_representation_property} and~\eqref{eqn_reproducing_property} is that the inner products reduce to arithmetic operations on point evaluations of $K(u,v)$. In particular, for arbitrary functions $f(x) = \sum_{v} \alpha_v K(x, v)$ and $g(x) = \sum_{u} \beta_u K(x, u)$ it holds that
\begin{equation}\label{eqn_inner_product}
   \big\langle 
              f(x), g(x) 
    \big\rangle_{\ccalH} 
                   =
                     \sum_{v,u} \alpha_v \overline{\beta}_u K(u, v)
      ,
\end{equation}
where $\overline{\beta}_{v}$ is the complex conjugate of $\beta_{v}$.

To ease notation and terminology, from now on, we use the term ``kernel" to refer to a positive semidefinite kernel.


\subsection{Classic Convolutions in RKHS through Integral Operators}
\label{sub_sec_rkhs_int_oper}

In some scenarios, the kernel $K$ associated with an RKHS can be linked to a linear integral operator with amenable properties that can be used to provide an additional characterization of the RKHS. To see this, and in the context of the notion introduced above, let us consider the operator $\boldsymbol{T}_{S}: L_{2}(\ccalX,\mu)\to L_{2}(\ccalX,\mu)$ given by
\begin{equation}\label{eq_sub_sec_rkhs_int_oper_1}
\boldsymbol{T}_{S} f
        =
        \int_{\ccalX}
             S(u,v)f(v)d\mu(v),
\end{equation}
where $S: \ccalX\times\ccalX \to \mbC$ is square integrable as a function of $v$ for any $u\in\ccalX$, $\mu$ is a measure on $\ccalX$, and $S(u,v)$ is called the \textit{symbol} of $\boldsymbol{T}_{S}$. It has been shown that the range of $\boldsymbol{T}_{S}$ on $L_{2}(\ccalX,\mu)$ determines an RKHS $\ccalH$ with kernel $K$ given by~\cite{paulsen2016introduction,wainwright2019high,ghojogh2021reproducing}
\begin{equation}\label{eq_kernel_from_symbols}
K(u,v)=
             \int_{\ccalX}
                   S(u,z)
                    \overline{S(v,z)}
                   d\mu (z)
                 ,
\end{equation}
and
\begin{equation}\label{eq_rkhs_range}
 \left\Vert 
      f
 \right\Vert_{\ccalH}
       =
       \inf 
           \left\lbrace 
                 \Vert g \Vert_{2}
                      :
                       f = \boldsymbol{T}_{S}g 
           \right\rbrace
           .
\end{equation}
This emphasizes that although $L_{2}(\ccalX,\mu)$ is not an RKHS, the subset of functions obtained as diffusions through $\boldsymbol{T}_{S}$ determine an RKHS whose kernel is obtained directly from $S(u,v)$ as in~\eqref{eq_kernel_from_symbols}. The operation performed in~\eqref{eq_kernel_from_symbols} is referred to as a \textit{box product}~\cite{paulsen2016introduction} or \textit{convolution}~\cite{halmos2012bounded}, and we can write~\eqref{eq_kernel_from_symbols} as $K(u,v)=\left(S\square S^{\ast}\right)(u,v)$ where $S^{\ast}(z,v)=\overline{S(v,z)}$. Additionally, with the appropriate choice of $S(u,v)$, one can represent classical convolutions as we discuss below.



\vspace{3mm}
\subsubsection{\underline{Convolutions on $\mbR$}}
\label{ex_sincaskernel} 

In~\eqref{eq_sub_sec_rkhs_int_oper_1}, one can choose $S(u,v)=h(u-v)$ on $\ccalX=\mbR$ with the Lebesgue measure to obtain
\begin{equation}
\boldsymbol{T}_{S} f
        =
          \int_{\mbR}   
                  h(u-v)
                  f(v)dv
         =
             h\star f         
                  .
\end{equation}
Then, we can see the classical ``$\star$" convolution of functions on $\mbR$ as the diffusion of a signal through the operator $\boldsymbol{T}_{S}$. This has the consequence that any function on $\mbR$ obtained as the result of the classic ``$\star$" convolution of two signals is in fact part of an RKHS, whose kernel is determined by a specific choice of $h$. One particularly well-known RKHS is obtained when 
\begin{equation}\label{eq_h_to_kernel}
h(u)=
         \frac{B}{\pi}\sinc\left(
                                          \frac{B}{\pi}u
                                    \right)
                                    .
\end{equation}
With this choice and $S(u,v)=h(u-v)$, we obtain through~\eqref{eq_kernel_from_symbols} the kernel 
\begin{equation}\label{eq:sinckit}
K(u,v)
      =
       \frac{B}{\pi}
           \sinc\left(
                   \frac{B}{\pi}
                       \left( 
                          u-v
                       \right)   
                 \right)
       ,
\end{equation}
which is the reproducing kernel of the RKHS consisting of the space of all the bandlimited signals with bandwidth $[-B,B]$ with the $L_{2}$ inner product $\langle f, g \rangle_{L_{2}}=\int_{-\infty}^{\infty}f\overline{g}dx$~\cite{rkhs_conv}. This comes naturally when recalling that the ``$\star$" convolution encapsulates the operation of \textit{filtering} and that the Fourier transform of the convolution of two signals equals the product of their Fourier transform. The filtering of an arbitrary signal $f$ with a $[-B,B]$-bandlimited signal $h$ as in~\eqref{eq_h_to_kernel} results in a signal with bandwidth $[-B,B]$.


\subsection{RKHS and Integral Operators with Continuous Kernels}
\label{sub_sec_rkhs_on_int_cont_K}

The relationship between the operator $\boldsymbol{T}_{S}$ and its induced RKHS becomes even more intricate when we choose $S$ to be a continuous kernel on a compact domain, i.e. $S=K$, where $K:\ccalX\times\ccalX\to\mbC$ is a continuous reproducing kernel and $\ccalX$ is compact. When this is the case, there is a countable collection of orthonormal continuous functions $\{ \vartheta_i \}_{i=1}^{\infty}$ in $L_{2}(\ccalX,\mu)$ that are eigenvectors for $\boldsymbol{T}_{K}$ with corresponding nonnegative eigenvalues $\{ \sigma_i \}_{i=1}^{\infty}$ such that for every $f\in L_{2}(\ccalX,\mu)$ we have
\begin{equation}\label{eq_TK_operator}
\boldsymbol{T}_{K}f
      =
      \sum_{i=1}^{\infty}
             \sigma_{i}
                  \left\langle
                        f
                        ,
                        \vartheta_i
                  \right\rangle_{L_{2}}
                  \vartheta_{i}
                  ,
\end{equation}
and
\begin{equation}\label{eq_mercer_K_decomp}
K(u,v)
     =
      \sum_{i=1}^{\infty}
            \sigma_{i}
                    \vartheta_{i}(u)
                    \overline{\vartheta_{i}(v)}
            ,
\end{equation}
where the sum in~\eqref{eq_mercer_K_decomp} is uniformly and absolutely convergent~\cite{paulsen2016introduction,wainwright2019high,ghojogh2021reproducing}. The decomposition of the kernel in~\eqref{eq_mercer_K_decomp} in terms of the eigenvectors of $\boldsymbol{T}_{K}$ allows a direct interpretation of the diffusion of  $f$ through $\boldsymbol{T}_{K}$. In fact, the sequence of coefficients of the sums in~\eqref{eq_TK_operator} and~\eqref{eq_mercer_K_decomp} can be used to compare the functions $g(u)=\left(\boldsymbol{T}_{K}f\right)(u)$ and $k_{v}(u)=K(u,v)$ as expansions on $\{ \vartheta_{i}(u) \}_{i=1}^{\infty}$. Such expansion can also be used to build a notion of the Fourier transform for those functions in the RKHS, as is the case in graphon signal processing, detailed in the following subsection.


\subsubsection{\underline{Convolutions on Graphons}}
\label{subsubsec_gphonsp}


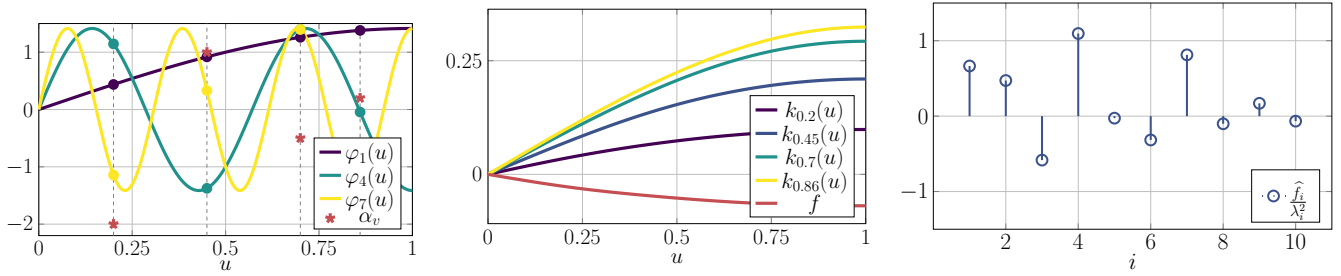
\begin{figure*}
%
%
       \begin{subfigure}{.32\linewidth}
          \resizebox{\textwidth}{!}{
\definecolor{viridis1}{RGB}{68,1,84}      
\definecolor{viridis2}{RGB}{33,145,140}   
\definecolor{viridis3}{RGB}{253,231,37}   
\definecolor{mygray}{RGB}{196,78,82}      
\definecolor{vertlinegray}{RGB}{128,128,128} 

\begin{tikzpicture}
	\begin{axis}[
		width=12cm,
		height=7.5cm,
		xlabel={$u$},
		ylabel={},
		xmin=0, xmax=1,
		ymin=-2.2, ymax=1.5,
		grid=major,
		xtick distance=0.25,        
		ytick distance=1,        
		legend pos=south east,
		legend style={font=\LARGE, draw=black, line width=0.4pt},
		legend image post style={line width=2.5pt},
		samples=200,
		line width=2.5pt,
		axis line style={line width=0.4pt},
		tick label style={font=\LARGE},
		label style={font=\LARGE},
		]
		\addplot[vertlinegray, dashed, line width=0.8pt, forget plot] 
		coordinates {(0.2, -2.2) (0.2, 1.5)};
		\addplot[vertlinegray, dashed, line width=0.8pt, forget plot] 
		coordinates {(0.45, -2.2) (0.45, 1.5)};
		\addplot[vertlinegray, dashed, line width=0.8pt, forget plot] 
		coordinates {(0.7, -2.2) (0.7, 1.5)};
		\addplot[vertlinegray, dashed, line width=0.8pt, forget plot] 
		coordinates {(0.86, -2.2) (0.86, 1.5)};
		\addplot[viridis1, domain=0:1] {sqrt(2)*sin(deg((1-0.5)*pi*x))};
		\addlegendentry{$\varphi_1(u)$}
		\addplot[viridis2, domain=0:1] {sqrt(2)*sin(deg((4-0.5)*pi*x))};
		\addlegendentry{$\varphi_4(u)$}
		\addplot[viridis3, domain=0:1] {sqrt(2)*sin(deg((7-0.5)*pi*x))};
		\addlegendentry{$\varphi_7(u)$}
		\addplot[viridis1, only marks, mark=*, mark size=3pt, forget plot] 
		coordinates {
			(0.2, {sqrt(2)*sin(deg((1-0.5)*pi*0.2))})
			(0.45, {sqrt(2)*sin(deg((1-0.5)*pi*0.45))})
			(0.7, {sqrt(2)*sin(deg((1-0.5)*pi*0.7))})
			(0.86, {sqrt(2)*sin(deg((1-0.5)*pi*0.86))})
		};
		\addplot[viridis2, only marks, mark=*, mark size=3pt, forget plot] 
		coordinates {
			(0.2, {sqrt(2)*sin(deg((4-0.5)*pi*0.2))})
			(0.45, {sqrt(2)*sin(deg((4-0.5)*pi*0.45))})
			(0.7, {sqrt(2)*sin(deg((4-0.5)*pi*0.7))})
			(0.86, {sqrt(2)*sin(deg((4-0.5)*pi*0.86))})
		};
		\addplot[viridis3, only marks, mark=*, mark size=3pt, forget plot] 
		coordinates {
			(0.2, {sqrt(2)*sin(deg((7-0.5)*pi*0.2))})
			(0.45, {sqrt(2)*sin(deg((7-0.5)*pi*0.45))})
			(0.7, {sqrt(2)*sin(deg((7-0.5)*pi*0.7))})
			(0.86, {sqrt(2)*sin(deg((7-0.5)*pi*0.86))})
		};
		\addplot[mygray, only marks, mark=star, mark size=4pt] 
		coordinates {
			(0.2, -2)
			(0.45, 1)
			(0.7, -0.5)
			(0.86, 0.2)
		};
		\addlegendentry{$\alpha_v$}
	\end{axis}
\end{tikzpicture}}
       \end{subfigure}
    	\begin{subfigure}{.32\linewidth}
		\centering
            \includegraphics[width=1\textwidth]{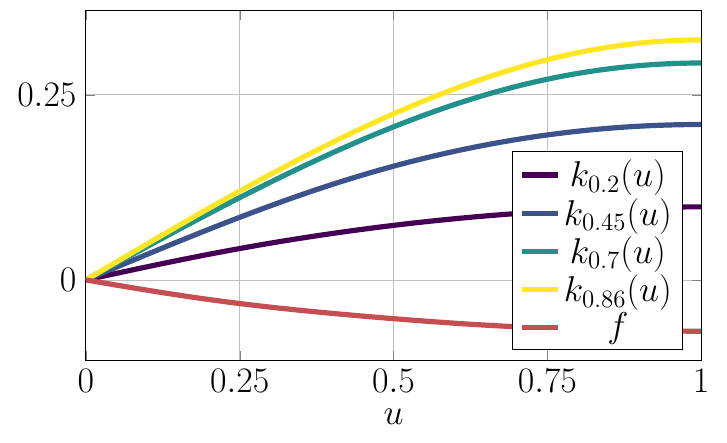} 
\end{subfigure}
       \begin{subfigure}{.32\linewidth}
          \resizebox{\textwidth}{!}{	\definecolor{stemblue}{RGB}{59,82,139}
	
	\begin{tikzpicture}
	\begin{axis}[
		width=12cm,
		height=7.5cm,
		xlabel={$i$},
		ylabel={},
		xmin=0, xmax=11,
		ymin=-1.5, ymax=1.5,
		grid=major,
		legend pos=south east,
		legend style={font=\LARGE, draw=black, line width=0.4pt, inner sep=7pt},
		legend image post style={line width=1.5pt},
		xtick={2,4,6,8,10},  
		axis line style={line width=0.4pt},
		tick label style={font=\LARGE},
		label style={font=\LARGE},
		]
		
		
		\addplot[
		stemblue, 
		ycomb,
		mark=o,
		mark size=4pt,
		line width=1.5pt,
		mark options={fill=stemblue, draw=stemblue}
		] coordinates {
			(1, 0.6648)
			(2, 0.4721)
			(3, -0.5831)
			(4, 1.0954)
			(5, -0.0268)
			(6, -0.3162)
			(7, 0.8126)
			(8, -0.1015)
			(9, 0.1697)
			(10, -0.0665)
		};
		\addlegendentry{$\frac{\widehat{f}_{i}}{\lambda_{i}^{2}}$}
		
	\end{axis}
\end{tikzpicture}}
       \end{subfigure}
%
%
\caption{Illustration for Example~\ref{exm_cor_gphon_fourier_kv}. Left: Discrete representation showing eigenfunctions $\varphi_{1}(u)$, $\varphi_{4}(u)$, and $\varphi_{7}(u)$ (curves) evaluated at sample points $u \in \{0.2, 0.45, 0.7, 0.86\}$ (markers), with coefficient vector $\boldsymbol{\alpha}=[-2,1,-0.5,0.2]$ indicated by $\alpha_v$. Center: Graphon signal $f(u)$ and kernel functions $k_{v}(u)$ for $v \in \{0.2, 0.45, 0.7, 0.86\}$ as functions of $u \in [0,1]$. Right: Normalized Fourier coefficients $\widehat{f}_{i}/\lambda_{i}^{2}$.}
    \label{fig_cor_gphon_fourier_kv}
\end{figure*}


A graphon is a symmetric measurable function on the unit square whose images lie on the interval $[0,1]$. Initially introduced to model large graphs and the limits of sequences of graphs, graphons have become essential tools for understanding how information diffuses in large networks~\cite{gphon_pooling_c,gphon_pooling_j,gphon_samp,lovaz2012large,diao2016model,gphon_leus}. The basic framework to understand such diffusions is known as \textit{graphon signal processing (Gphon-SP)}, and it is a particular instantiation of an algebraic signal model as in~\cite{algSP0,algSP1,algSP2,parada_algnn,algnn_nc_j,parada_algnnconf}. In Gphon-SP, signals are functions in $L_{2}[0,1]$ and filters are polynomials where the independent variable is the so-called graphon shift operator $\boldsymbol{T}_{W}$ given by
\begin{equation}\label{eq_Tw}
\boldsymbol{T}_{W}
        f
        =
        \int_{0}^{1}
             W(u,v)f(v)dv
             ,
\end{equation}
where $W(u,v)$ is the graphon. The convolution between a signal $f\in L_{2}[0,1]$ and a filter $p(\boldsymbol{T}_{W})=\sum_{r=0}^{R-1}h_{r}\boldsymbol{T}_{W}^{r}$ is given by
\begin{equation}
p\left(
        \boldsymbol{T}_{W}
   \right)
        f
        =
       \sum_{r=0}^{R-1}h_{r}\boldsymbol{T}_{W}^{r}f
             ,
\end{equation}
where $\boldsymbol{T}_{W}^{r}$ indicates the $r$-times application of $\boldsymbol{T}_{W}$. The boundedness of $W(u,v)$ and the compactness of $[0,1]$ ensures that $\boldsymbol{T}_{W}$ is Hilbert-Schimidt, which endows $\boldsymbol{T}_{W}$ with a collection of eigenvectors $\{ \varphi_{i}(u) \}_{i=1}^{\infty}$ and eigenvalues $\{ \lambda_{i} \}_{i=1}^{\infty}$ such that
\begin{equation}\label{eq_Tw_eigendec}
\boldsymbol{T}_{W}f
      =
      \sum_{i=1}^{\infty}
             \lambda_{i}
                  \left\langle
                        f
                        ,
                        \varphi_i
                  \right\rangle_{L_{2}}
                  \varphi_{i}
                  ,
\end{equation}
and
\begin{equation}\label{eq_W_eigendec}
W(u,v)
     =
      \sum_{i=1}^{\infty}
            \lambda_{i}
                    \varphi_{i}(u)
                    \varphi_{i}(v)
            ,
\end{equation}
where the sum in~\eqref{eq_W_eigendec} is $L_{2}$-convergent~\cite{lovaz2012large}. Despite the uncanny resemblance between~\eqref{eq_Tw_eigendec},~\eqref{eq_W_eigendec} and~\eqref{eq_TK_operator},~\eqref{eq_mercer_K_decomp} there is a fundamental difference,  $\sigma_{i}\geq 0$, while 
$\lambda_{i}\in[-1,1]$ for all $i$. At the same time, there is a direct relationship between the eigendecompositions in both scenarios. Due to the importance of such a connection, we make this clear in the following result.


\begin{theorem}
\label{thm_K_from_W_gphonsp}
Let $W(u,v)$ be a graphon with graphon shift operator $\boldsymbol{T}_{W}$, whose eigenvectors and eigenvalues are $\{ \varphi_{i}(u) \}_{i=1}^{\infty}$ and $\{ \lambda_{i} \}_{i=1}^{\infty}$, and let
\begin{equation}
\ccalH 
        =
          \left\lbrace
                \left. 
                      \boldsymbol{T}_{W}f
                \right\vert
                             f\in L_{2}[0,1]      
          \right\rbrace
          .
\end{equation}
Then, $\ccalH$ is an RKHS with reproducing kernel $K=W\square W$. Additionally, if $\{ \vartheta_{i} \}_{i=1}^{\infty}$ and $\{ \sigma_{i} \}_{i=1}^{\infty}$ are the eigenvectors and eigenvalues of $\boldsymbol{T}_{K}$, respectively, it follows that
\begin{equation}\label{eq_thm_K_from_W_gphonsp_2}
\boldsymbol{T}_{K}
              =
                \boldsymbol{T}_{W}^{2}
                ,
\end{equation}
\begin{equation}
 \vartheta_{i}(u)
       =
        \varphi_{i}(u),\quad\forall i\in\mbN,~\forall u\in [0,1]
        ,
\end{equation}
and
\begin{equation}\label{eq_thm_K_from_W_gphonsp_3}
\sigma_{i}
       =
        \lambda_{i}^{2}
        ,
        \quad i\in\mbN
        .
\end{equation}
\end{theorem}

\begin{proof}
See Appendix~\ref{proof_thm_K_from_W_gphonsp}.
\end{proof}


Although there is a clear distinction between graphons and their induced kernels, Theorem~\ref{thm_K_from_W_gphonsp} emphasizes in~\eqref{eq_thm_K_from_W_gphonsp_2} that any polynomial diffusion carried out with $\boldsymbol{T}_{K}$ is equivalent to a diffusion with $\boldsymbol{T}_{W}$ with an even polynomial, i.e. $p\left(\boldsymbol{T}_{K}\right)=p(\boldsymbol{T}_{W}^{2})$ for any polynomial $p$. One natural question that arises, given the strong relationship between graphons and RKHS stated in Theorem~\ref{thm_K_from_W_gphonsp}, is whether a Kernel can naturally induce a graphon. We provide an answer to this question in the following result.


\begin{theorem}\label{thm_W_gphon_from_K}
Let $K: [0,1]^2 \to \mbR$ be a continuous kernel with $K(u,v)\geq 0$ on $[0,1]^2$. Then, there is an induced graphon $W$ on $[0,1]^2$ given by
\begin{equation}\label{eq_thm_W_gphon_from_K_1}
W(u,v)=
       \frac{1}{C}
            K(u,v)
            ,
\end{equation}
where
\begin{equation}\label{eq_thm_W_gphon_from_K_2}
C = 
     \sup_{(u,v)\in [0,1]^2}
               K(u,v)
                  .
\end{equation}
\end{theorem}

\begin{proof}
See Appendix~\ref{proof_thm_W_gphon_from_K}.
\end{proof}


\noindent In the following examples we derive the graphon associated to a collection of specific reproducing kernels, relying on Theorem~\ref{thm_W_gphon_from_K}.


\begin{example}\normalfont 
In the light of Theorem~\ref{thm_W_gphon_from_K} the following are examples of graphons induced by reproducing kernels:
\begin{itemize}
\item $K: [0,1]^{2}\to \mbR$ given by $K(u,v)=uv$ induces a graphon $W(u,v)=uv$.
\item The reproducing kernels $K(u,v)=\exp\left(-\frac{\vert u-v\vert}{\sigma} \right)$, $ K(u,v)=\exp\left(-\frac{2}{\ell^{2}}\sin^{2}\left( \pi\left(u-v\right)\right)\right)$ and $K(u,v)=\exp\left(-(u-v)^{2}/(2\sigma^2) \right)$ on $[0,1]^{2}$ induce graphons given by $W(u,v)=K(u,v)$.
\item On $[0,1]^2$ the reproducing kernel $K(u,v)=(1+uv)^2$ induces the graphon $W(u,v) =(1/4)\left(1+uv\right)^2$, i.e. $C=4$ in~\eqref{eq_thm_W_gphon_from_K_2}.
\end{itemize}
\end{example}





From the combined insights in Theorems~\ref{thm_K_from_W_gphonsp} and~\ref{thm_W_gphon_from_K}, we have that the graphon Fourier transform is always linked to an RKHS representation. Let us recall that when a graphon shift operator $\boldsymbol{T}_{W}$ has eigenvectors $\{ \varphi_{i}(u) \}_{i=1}^{\infty}$, we obtain the $i$-th graphon Fourier coefficient of $f$ as $\widehat{f}_{i}=\left\langle f, \varphi \right\rangle_{L_{2}}$. Then, the action of $\boldsymbol{T}_{K}$ on $f$ can be interpreted as the spectral filtering on $\widehat{f}$ where each Fourier coefficient is modulated by an eigenvalue $\sigma_{i}$. Additionally, the decomposition of the kernel in terms of the eigenvectors, combined with the representation property of the RKHS, allows us to interpret $k_{v}(u)=K(u,v)$ in terms of a special Fourier decomposition that we present in the following theorem.


\begin{theorem}\label{thm_gphon_fourier_kv}
Let $W$ be a graphon and let $K=W\square W$ be the reproducing kernel of an RKHS $\ccalH$ on $[0,1]$. If the eigenvectors and eigenvalues of $\boldsymbol{T}_{W}$ are $\{ \varphi_{i}(u) \}_{i=1}^{\infty}$ and $\{ \lambda_{i} \}_{i=1}^{\infty}$, respectively, then, the graphon Fourier transform coefficients of $k_{v}(u)=K(u,v)$ are given by
\begin{equation}\label{eq_thm_gphon_fourier_kv_1}
\left( 
      \widehat{k}_{v}
\right)_{i}
              =
                 \lambda_{i}^{2}
                 \varphi_{i}(v)
                           \quad
                                   \forall i\in\mbN
                 .
\end{equation}
Additionally, any $f\in\ccalH$ represented by $f=\sum_{v\in [0,1]}\alpha_{v}k_{v}(u)$ has graphon Fourier coefficients given by
\begin{equation}\label{eq_thm_gphon_fourier_kv_2}
\widehat{f}_{i}
              =
              \sum_{v\in [0,1]}
                   \alpha_{v}
                   \left( 
                        \widehat{k}_{v}
                   \right)_{i}
              =
                 \lambda_{i}^{2}
                       \sum_{v\in [0,1]}
                                \alpha_{v}
                                \varphi_{i}(v)
                           \quad
                                  \forall i\in\mbN
                 ,
\end{equation}
where $\alpha_{v}\in\mbC$.
\end{theorem}

\begin{proof}
    See Appendix~\ref{proof_thm_gphon_fourier_kv}
\end{proof}


The result in Theorem~\ref{thm_gphon_fourier_kv}, although simple, has a profound implication on how one interprets the functions $k_{v}(u)$ on the spectral domain. In particular, one can see that when using the representation property to expand an arbitrary function in $\ccalH$, its Fourier coefficients are obtained as an inner product modulated by an eigenvalue, i.e., if we consider that $\boldsymbol{\alpha}=[\alpha_{v}]_{v\in\ccalV}$ and $\boldsymbol{\varphi}_{i}=[\varphi_{i}(v)]_{v\in\ccalV}$ are infinite dimensional vectors indexed by $v\in\ccalV$, then, we have $\widehat{f}_{i}=\lambda_{i}^{2}\boldsymbol{\alpha}^{T}\boldsymbol{\varphi}_{i}$. In this context, we can see the graphon Fourier representation of $f$ as determined by a discrete spectrum obtained from sampling the Fourier domain precisely on those values specified by $v\in\ccalV$.


\begin{example}\normalfont 
\label{exm_cor_gphon_fourier_kv}
To better understand Theorem~\ref{thm_gphon_fourier_kv}, let us consider the case where $W=\min(u,v)$. Such graphon has an associated graphon operator $\boldsymbol{T}_{W}$ with eigenfuctions and eigenvalues given by
\begin{equation}
\varphi_{i}(u)
   =
    \sqrt{2}
    \sin\left( 
              \left(
                  i - \frac{1}{2}
               \right)
               \pi u
        \right)
        ,
        \quad 
        \lambda_{i}=
        \frac{1}
        {\left(
               i-\frac{1}{2}
         \right)^{2}\pi^2} 
         ,
\end{equation}
for $i=1,2,\ldots$. By Theorem~\ref{thm_K_from_W_gphonsp} we obtain a reproducing kernel $K=W\square W$ with eigenvectors $\varphi_{i}(u)$ and eigenvalues $\lambda_{i}^{2}$. Then, in the light of~\eqref{eq_mercer_K_decomp} we have that
$
k_{v}(u)
      =
       \sum_{i=1}^{\infty}
            \lambda_{i}^{2}
                    \varphi_{i}(u)
                    \varphi_{i}(v)
$. Now, let us consider the signal in the RKHS associated to $K$ and given by
\begin{equation}
f(u)=-2k_{0.2}(u)+k_{0.45}(u)-0.5k_{0.7}(u)+0.2k_{0.86}(u)    
.
\end{equation}
Equation~\eqref{eq_thm_gphon_fourier_kv_2} in Corollary~\ref{thm_gphon_fourier_kv} establishes that the $i$-th Fourier coefficient can be computed via a finite discrete inner product. Specifically, we take the inner product between the representation coefficients of $f$, given by $\boldsymbol{\alpha}=[-2,1,-0.5,0.2]$, and the samples of the eigenfunction $\varphi_{i}(u)$ evaluated at the centers of the kernel functions $k_{v}(u)$, yielding $\boldsymbol{\varphi}_{i}=[\varphi_{i}(0.2), \varphi_{i}(0.45), \varphi_{i}(0.7), \varphi_{i}(0.86)]$. Thus, each Fourier coefficient of $f$ can be obtained as
\begin{equation}
\widehat{f}_{i}
    =
     \lambda_{i}^{2}
     \left( 
          -\varphi_{i}(0.2)
          +\varphi_{i}(0.45)
          -0.5\varphi_{i}(0.7)
          +0.2\varphi_{i}(0.86)
     \right)
     .
\end{equation}
Figure~\ref{fig_cor_gphon_fourier_kv} provides a geometric illustration on how $\widehat{f}_{i}$ is related to $\boldsymbol{\alpha}$ and $\boldsymbol{\varphi}_{i}$.
\end{example}



\begin{remark}\normalfont
Notice that when discussing the notion of the Fourier transform on graphons, we consider the eigenvectors $\{ \varphi_{i}\}_{i=1}^{\infty}$ as an orthonormal basis in $L_{2}[0,1]$. Then, the $i$-th Fourier coefficient of any $f\in L_{2}[0,1]$ is determined by $\widehat{f}_{i}=\left\langle f,\varphi_{i} \right\rangle_{L_{2}}$. However, it is possible to define a notion of Fourier transform on the RKHS, $\ccalH$, induced by the graphon. In such a case, a scaling is necessary since the $\{ \varphi_{i}\}_{i=1}^{\infty}$ are orthogonal but not orthonormal in the RKHS. More specifically, we have that $\left\langle \varphi_{i}, \varphi_{i}\right\rangle_{\ccalH}=1/\sigma_{i}$. This means that while $\{ \varphi_{i}\}_{i=1}^{\infty}$ is the orthonormal basis in $L_{2}[0,1]$, $\{ \sqrt{\sigma_{i}}\varphi_{i}\}_{i=1}^{\infty}$ is the orthonormal basis in $\ccalH$.
\end{remark}



\subsubsection{\underline{Implications on Gphon-SP with Digraphons}}
Graphon signal processing focuses on information processing over large \textit{undirected} graphs. This restriction stems from the symmetry condition $W(u,v)=W(v,u)$, which is fundamental to the definition of a graphon. While this symmetry is necessary to ensure the self-adjointness and favorable spectral properties of $\boldsymbol{T}_{W}$~\cite{lovaz2012large,digraphon1,digraphon2,digraphon3}, it is not required for the notion of filtering itself---polynomial filters can be properly defined even with non-self-adjoint shift operators~\cite{algSP0,algnn_nc_j,parada_algnn}. The key advantage of self-adjointness is that a polynomial function of $\boldsymbol{T}_{W}$ reduces to the same polynomial function of its eigenvalues in~\eqref{eq_Tw_eigendec}. This property enables direct interpretability of filters through their effect on the Fourier representation---a feature absent for non-self-adjoint operators.

To address the spectral challenges posed by non-symmetric graphons $W:[0,1]^{2}\to [0,1]$, we exploit the RKHS induced by $\boldsymbol{T}_{W}$ through the continuous kernel $K$ in~\eqref{eq_mercer_K_decomp}, expressed in terms of an orthonormal basis for $L_{2}[0,1]$. Using this basis, we demonstrate that a subclass of polynomial diffusions in the node domain can be recast as pointwise polynomial modulations of the Fourier coefficients. We formalize these ideas in the following result, where we refer to any bounded measurable function $W:[0,1]^{2}\to [0,1]$ as a \textit{digraphon}---note that we do not require $W(u,v)=W(v,u)$.


\begin{theorem}\label{thm_diff_digraphon}
Let $W: [0,1]^{2}\to [0,1]$ be a digraphon and let $\boldsymbol{T}_{W}$ be the operator defined in~\eqref{eq_Tw}. Define the range space
$
\mathcal{H} 
        =
          \left\lbrace
                \boldsymbol{T}_{W}f
                \,:\,
                f\in L_{2}[0,1]      
          \right\rbrace.
$
Then $\mathcal{H}$ is a reproducing kernel Hilbert space (RKHS) with reproducing kernel $K=W\square W$. Moreover, if $K$ is nontrivial, then $\boldsymbol{T}_{K}$ and $K$ admit the spectral representations given in~\eqref{eq_TK_operator} and~\eqref{eq_mercer_K_decomp}, respectively. Furthermore, the operator $\boldsymbol{T}_{K}$ satisfies
\begin{equation*}
\boldsymbol{T}_{K}=\boldsymbol{T}_{W}\boldsymbol{T}_{W^{\ast}},
\end{equation*}
where $\boldsymbol{T}_{W^{\ast}}$ is given by
\begin{equation}
 \boldsymbol{T}_{W^{\ast}}f
       =
        \int_{0}^{1}W(u,v)f(u)\,du.
\end{equation}
\end{theorem}

\begin{proof}
See Appendix~\ref{proof_thm_diff_digraphon}.
\end{proof}


A direct insight from Theorem~\ref{thm_diff_digraphon} is that we can use $\{ \vartheta_{i} \}_{i=1}^{\infty}$  to define a notion of Fourier transform for any signal on the digraphon. Naturally, the usefulness of $\boldsymbol{T}_{K}$ relies on $W\square W$ not resulting on the trivial kernel. Finally, notice that the analysis on digraphons serves directly as a signal processing model for digraphs. This is a consequence of the one-to-one relationship between adjacency matrices and their induced representations as functions on $[0,1]^2$~\cite{lovaz2012large}, which also causes a one-to-one relationship between signals on directed graphs and signals on digraphons~\cite{diao2016model,gphon_samp,gphon_pooling_j}.




\section{Algebraic Convolutions with Integral Operators in RKHS}
\label{sec_alg_conv_filt_int_opt_rkhs}

The notions of filtering, polynomial diffusions, and more generally signal models can be described in the framework of \textit{algebraic signal processing (ASP)}, which has emerged as a fundamental theory to represent consistently a wide diversity of convolutional signal frameworks such as discrete-time signal processing~\cite{algSP1}, discrete space models with symmetric shift operators~\cite{algSP2}, signal models on 2D hexagonal lattices~\cite{algSP6}, signal models on general lattices~\cite{puschel_asplattice}, signal processing on sets~\cite{puschel_aspsets}, quiver signal processing~\cite{parada_quiversp}, Lie group signal processing~\cite{lga_j,lga_icassp}, graphon signal processing~\cite{gphon_pooling_c,gphon_pooling_j,gphon_samp,lovaz2012large,diao2016model,gphon_leus}, multigraph signal processing~\cite{msp_j,msp_icassp2023}, signal models on digraphs~\cite{puschel_digraphs1,puschel_digraphs2}, among others~\cite{algSP7,algSP8,algnn_nc_j}.

In ASP, an algebraic signal model (ASM) is determined by the triplet 
\begin{equation}
\left( 
       \ccalA
       ,
       \ccalH
       ,
       \rho
\right)
,
\end{equation}
where $\ccalA$ is an associative unital algebra, $\ccalH$ is a vector space, and $\rho: \ccalA \to\text{End}(\ccalH)$ is a homomorphism from $\ccalA$ in to the space of linear maps from $\ccalH$ onto itself, $\text{End}(\ccalH)$.

The algebra, $\ccalA$, is a vector space with a closed operation of product. The quintessential example of an algebra is the space of polynomials with coefficients in $\mbC$, $\mbC[t]$. It is well known that $\mbC[t]$ is a vector space with the ordinary operations of sum and multiplication by scalars. What turns $\mbC[t]$ into an algebra is an operation of product, which in this case is the ordinary multiplication between polynomials. Such a product is closed since the ordinary multiplication of two polynomials results in a polynomial as well. Additionally, $\mbC[t]$ is a unital algebra as the degree zero polynomial, $p(t)=1$, becomes an identity concerning the product defined in $\mbC[t]$. We refer to the elements of $\ccalA$ as the filters in the ASM. As discussed in~\cite{algSP0,parada_algnn,algnn_nc_j,parada_algnnconf}, the notion of an algebra is very rich and extends beyond the example just discussed. Moreover in~\cite{rkhs_conv} it is shown that RKHS naturally induce algebras associated to the domain of the reproducing kernel.

The vector space, $\ccalH$, contains the information that we want to process with the given signal model. This is where, with $\ccalH$, we endow data with an algebraic structure. We refer to the elements in $\ccalH$ as the \textit{signals}.

The homomorphism, $\rho$, is a linear map from $\ccalA$ to the space of linear maps from $\ccalH$ onto itself, $\text{End}(\ccalH)$, that preserves the product in $\ccalA$. This means that for any $a,b\in\ccalA$ it must follow that
\begin{equation}\label{eq_hom_def}
\rho\left( 
             ab
      \right)
              =
                \rho(a)
                \rho(b)
                ,
\end{equation}
where $\rho(a), \rho(b)\in\text{End}(\ccalH)$ are linear operators. Then, under~\eqref{eq_hom_def} we can conceive $\rho$ as the algebraic object that physically implements the abstract filters in $\ccalA$ into concrete operators that transform the data in $\ccalH$.


\subsection{Homomorphisms with Integral Operators}

The implementation of filters in some ASMs is carried out using integral operators. This is the case in Gphon-SP, where the ordinary polynomials are turned into polynomial operators where the independent variable is $\boldsymbol{T}_{W}$. More specifically, Gphon-SP can be considered a particular instantiation of the generic ASM $\left(\mbC[t],L_{2}(\ccalX,\mu),\rho_{S}\right)$, where $\mbC[t]$ is the algebra of polynomials with one independent variable and coefficients in $\mbC$, $L_{2}(\ccalX,\mu)$ is the space of square integral functions on $\ccalX$ with the measure $\mu$, and $\rho_{S}: \mbC[t]\to \text{End}\left(L_{2}(\ccalX,\mu)\right)$ is the homomorphism given by $\rho_{S}(t)=\boldsymbol{T}_{S}$, with $\boldsymbol{T}_{S}$ given by~\eqref{eq_sub_sec_rkhs_int_oper_1}, and where $S: \ccalX\times\ccalX\to\mbC$ is an $L_{2}$-integrable function on $\ccalX\times\ccalX$. Since $\rho_{S}$ is a homomorphism, it is a linear map that preserves the products in $\mbC[t]$, which guarantees that for any polynomial $p(t)\in\mbC[t]$ we must have
\begin{equation}
\rho_{S}\left(
              p(t)
       \right)
               =
                 p\left( 
                          \rho_{S}(t)
                   \right)
                            =
                                p\left( 
                                       \boldsymbol{T}_{S}
                                  \right)
                                  .
\end{equation}

Then, under the light of the discussion in Section~\ref{sub_sec_rkhs_int_oper}, we know that there is a subclass of signals $\ccalH\subset L_{2}(\ccalX,\mu)$ that constitute an RKHS with a reproducing kernel given by $K(u,v)=\left(S\square S^{*}\right)(u,v)$. Additionally, when choosing $S=K$, where $K$ is a kernel inducing the RKHS $\ccalH(K)$, one can focus on processing information with the ASM $\left(\mbC[t],\ccalH(K),\rho_{K}\right)$.

From an algebraic point of view, $\left(\mbC[t],L_{2}(\ccalX,\mu),\rho_{S}\right)$ and $\left(\mbC[t],\ccalH(K),\rho_{K}\right)$ embed the attributes of an RKHS in the space of signals. This is somewhat natural since the RKHS is a vector space. Additionally, notice that while it is easier to leverage the representation and reproducing properties of the RKHS in $\left(\mbC[t],\ccalH(K),\rho_{K}\right)$, the model $\left(\mbC[t],L_{2}(\ccalX,\mu),\rho_{S}\right)$ allows for a more extense family of functions to model the datasets.





\subsection{Homomorphisms with RKHS Algebras}
\label{sub_sec_hom_rkhs_alg}


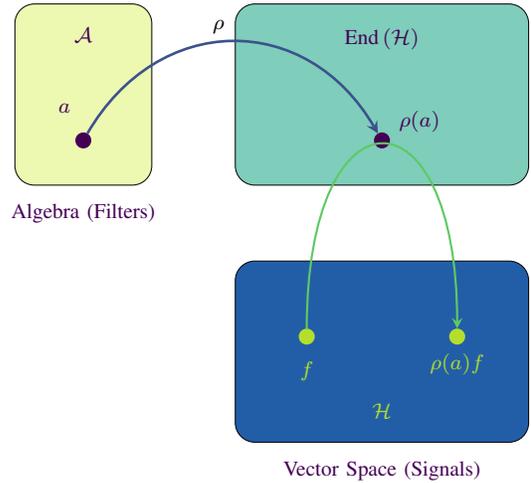
\begin{figure}
        \centering


\definecolor{my_cp_col1}{RGB}{253, 231, 37}
\definecolor{my_cp_col2}{RGB}{180, 222,44}
\definecolor{my_cp_col3}{RGB}{94, 201, 98}
\definecolor{my_cp_col4}{RGB}{33, 145, 140}
\definecolor{my_cp_col5}{RGB}{59, 82, 139}
\definecolor{my_cp_col6}{RGB}{68, 1, 84}

\definecolor{my_cp2_col1}{RGB}{255,255,217}
\definecolor{my_cp2_col2}{RGB}{237,248,177}
\definecolor{my_cp2_col3}{RGB}{199,233,180}
\definecolor{my_cp2_col4}{RGB}{127,205,187}
\definecolor{my_cp2_col5}{RGB}{65,182,196}
\definecolor{my_cp2_col6}{RGB}{29,145,192}
\definecolor{my_cp2_col7}{RGB}{34,94,168}
\definecolor{my_cp2_col8}{RGB}{37,52,148}
\definecolor{my_cp2_col9}{RGB}{8,29,88}

\usetikzlibrary{positioning,decorations.pathreplacing,shapes}


\def \scale {1.3}
\def \unit{\scale cm}
\def \layerinterdist {4}

\tikzstyle{set} = [rectangle,color=black,
                    rounded corners = 0.2*\unit,
                    fill=black,
                    inner sep=0pt,
                    draw,
                    anchor = center,
                    line width=0.1mm]
                                      
\tikzstyle{vectorspace} = [set, 
                             fill=my_cp2_col7,
                             minimum width  = 3*\unit,
                             minimum height = 1.8541*\unit]
                             
\tikzstyle{endomorphisms} = [vectorspace,
                              fill=my_cp2_col4]
                                
\tikzstyle{algebra} = [endomorphisms,
                        fill=my_cp2_col2,
                         minimum width  = 1.4*\unit,
                        minimum height = 1.8541*\unit]


\tikzstyle{dot} = [circle,
                    minimum width  = 0.12*\unit,
                    fill=black,
                    color=black,
                    inner sep=0pt,
                    draw,
                    anchor = center ]

{\fontsize{8}{8}\selectfont

\begin{tikzpicture}[rounded corners,ultra thick]


   \path (0,0) node [vectorspace, fill opacity=1] (M0) {};
   \path (M0.south) ++ (0, 0.2) node [above, color=my_cp_col2] {$\mathcal{H}$};
   
   \path (M0) ++ (-1,0.2) node [dot,color=my_cp_col2] (x) {};
   \path (x.south)++(0,-0.1) node [below, color=my_cp_col2] {$f$}; 
   
   \path (M0) ++ (1,0.2) node [dot,color=my_cp_col2] (ex) {};
   \path (ex.south) node [below, color=my_cp_col2] {$\rho(a)f$};

 \path (M0.south) ++ (0,0) node [] (Htext) {};
 \path (Htext.south) node [below, color=my_cp_col6,align=center] {Vector Space
   (Signals)};


   \path (M0.north) ++ (0, 1) 
         node [endomorphisms, anchor=south, fill opacity=1] (End0) {};
   \path (End0.north) ++ (0.0, -0.2) node [below, color=my_cp_col6] {$\text{End}\left(\mathcal{H}\right)$};

   \path (End0.south) ++ (0.0, 0.6) node [dot,color=my_cp_col6] (e) {};      
   \path (e.center) ++ (0.5,0.25) node [color=my_cp_col6] {$\rho(a)$}; 
   
   \path (e)+(-1,0.77) coordinate (c1);
   \path (e)+(1,0.77) coordinate (c2);   
   \path [draw, -stealth, line width=0.8,color=my_cp_col3] (x) .. controls (c1) and (c2) .. (ex);


   \path (End0.north west) ++ (-1.1, 0) 
         node [algebra, anchor = north east, fill opacity=1] (A) {};   
   \path (A.north) ++ (0,-0.2) node [below,color=my_cp_col6] {$\mathcal{A}$};
   \path (A.center) ++ (-0.25,0) node [below,color=my_cp_col6] {$a$};  
    
    
    \path (A.south) ++ (0.0, 0.6) node [dot,color=my_cp_col6] (a1) {}; 
        
   \path (A.north) + (1,-0.1) coordinate (c1);
   \path (End0.north) + (-1,-0.1) coordinate (c2);   
   \path [draw, -stealth,line width = 1,color=my_cp_col5,  opacity=1] (a1) .. controls (c1) and (c2) .. (e) node[midway,above left,rotate=0,color=black,opacity=1] {$\rho$};

 \path (A.south) ++ (0,0) node [] (Atext) {};
 \path (Atext.south) node [below, color=my_cp_col6,align=center] {Algebra
   (Filters)};

%
%
%
%
%

\end{tikzpicture}
} 
        \caption{Pictorial representation of a generic algebraic signal model (ASM) $(\ccalA, \ccalM, \rho)$. Filters are elements of the algebra $\ccalA$ while the signals are the elements of the vector space $\ccalH$. The homomorphism, $\rho$, translates the abstract filters in $\ccalA$ into concrete operators in $\text{End}(\ccalH)$, that act on the signals in $\ccalH$. The symbol $\text{End}(\ccalH)$ represents the space of linear operators from $\ccalH$ onto itself.}
        \label{fig_asp_model}
\end{figure}


In the previous subsection, we emphasized how an RKHS can be encapsulated in an ASM. While this is somewhat natural, it is not the only way in which an RKHS is linked to an ASM. In what follows, we show that the reproducing kernel of an RKHS induces an algebra that can be used to incorporate the reproducing property of the RKHS in a homomorphism.

Let $K: \ccalX\times\ccalX\to \mbC$ be the reproducing kernel of the RKHS $\ccalH(K)$. Then, for $K$, it is possible to perform the $r$-times box product of $K$ with itself, which we represent by
\begin{equation}\label{eq_K_pow_box_r}
K^{\square r}
            =
                \underbrace{K\square\ldots\square\ldots\square K}_{r-\text{times}}
                ,
\end{equation}
where the measurability of $K$ guarantees that~\eqref{eq_K_pow_box_r} is always well defined. Notice that in virtue of~\eqref{eq_kernel_from_symbols} $K^{\square r}$ is a reproducing kernel on its own.

With~\eqref{eq_K_pow_box_r}, we can construct polynomials that naturally lead to well-defined functions on $\mathcal{X}\times\mathcal {X}$. In particular, a polynomial with coefficients $\{ a_{r} \}_{r=1}^{R}$ can be writen as
\begin{equation}
p\left(
        K^{\square}
  \right)
           =
              \sum_{r=1}^{R}
                             a_{r}
                             K^{\square r}
                             ,
\end{equation}
where $K^{\square}\equiv K^{\square 1}\equiv K$. With the notation at hand, we now show that polynomials based on the box product can be endowed with the structure of an algebra. This, beyond any formality and association with ASMs, will enable us to guarantee that products between polynomials and an identity element are well-defined.


\begin{theorem}
\label{thm_alg_RKHS_box_prod_1}
Let $K$ be a reproducing kernel on $\ccalX\times\ccalX$, $\mu$ be a finite Borel measure on $\ccalX$, and $\ccalA_{K}$ be the set given by
\begin{equation}\label{eq_A_rkhs_box_1}
\ccalA_{K} 
   =
   \left\lbrace 
      \left.
          \sum_{r=0}^{R}a_{r}K^{\square r}
      \right\vert 
          a_{r}\in\mbC,
          ~R\in\mbN
   \right\rbrace
   ,
\end{equation}
with $K^{\square 0}(u,v)=\delta(u-v)$, where $\delta$ is the delta dirac function. Additionally, let $\bullet: \ccalA_{K} \times \ccalA_{K} \to \ccalA_{K}$ be given by
\begin{equation}\label{eq_A_rkhs_box_2}
\left(
    \sum_{r=0}^{R}a_{r}K^{\square r}
\right)
     \bullet
\left(
    \sum_{\ell=0}^{L}b_{\ell}K^{\square\ell}
\right)  
     =
     \sum_{r=0, \ell=0}^{R,L}
           a_{r}b_{\ell}
           K^{\square (r+\ell)}
           .
\end{equation}
Then, $\ccalA_{K}$ with the product $\bullet$ is a unital algebra with identity element 
$
1_{\ccalA_{K}}
         =
            K^{\square 0}
                  ,
$
and where the addition and scalar multiplication in $\ccalA_{K}$ are given by
\begin{equation}\label{eq_A_rkhs_box_3}
\sum_{r=0}^{R}a_{r}K^{\square r}
                +
\sum_{r=0}^{R}b_{r}K^{\square r}    
                         =
                           \sum_{r=0}^{R}\left( 
                                                          b_{r}+a_{r}
                                                    \right)      
                                                          K^{\square r}  
                                                          ,
\end{equation}
and
\begin{equation}\label{eq_A_rkhs_box_4}
\alpha\left( 
                 \sum_{r=0}^{R}a_{r}K^{\square r}
          \right)
                       =
                           \sum_{r=0}^{R}
                                                    \left(
                                                           \alpha a_{r}   
                                                    \right)         
                                                          K^{\square r}  
                                                          ,
\end{equation}
for all $\alpha\in\mbC$.
\end{theorem}

\begin{proof} 
          See Appendix~\ref{proof_thm_alg_RKHS_box_prod_1}.
 \end{proof}


\noindent Notice that the box product with $K^{\square 0}$ is well defined when considering the basic properties of the impulse function, i.e. 
\begin{multline}
\left(
      K\square K^{\square 0}
\right)(u,v)      
           =
              \int_{\ccalX}
                     K(u,z)
                     \delta(z-v)
                     d\mu (z)
                     =
                     \\    
               \int_{\ccalX}
                      K(u,v)
                      \delta(z-v)
                      d\mu(z)
                     =
                       K(u,v)
                       .
\end{multline}

Theorem~\ref{thm_alg_RKHS_box_prod_1} provides the theoretical guarantee that the elements in $\ccalA_{K}$ can be treated \textit{like} ordinary polynomials, keeping in mind the specific product, sum, and multiplication by scalars stated in~\eqref{eq_A_rkhs_box_2},~\eqref{eq_A_rkhs_box_3}, and~\eqref{eq_A_rkhs_box_4}, respectively. 

Aligned with the properties of $\ccalA_{K}$, it is essential to emphasize the relationship between the algebraic powers on the box product and the spectral representation associated with $\boldsymbol{T}_{K}$. We describe such a relationship in the following result.


\begin{theorem}\label{thm_box_prod_spect}
Let $K$ be a continuous kernel on $\ccalX\times\ccalX$, where $\ccalX$ is a compact subset of $\mbR^{d}$ and let $\mu$ be a finite Borel measure on $\ccalX$. Let $\vartheta_i$ and $\sigma_i$ the eigenvectors and eigenvalues of the integral operator $\boldsymbol{T}_{K}: L^{2}(\ccalT,\mu)\to L^{2}(\ccalT,\mu)$, respectively, and
$
S_{1}(u,v) = 
        \sum_{i=1}^{\infty}
             a_{i}
             \vartheta_{i}(u)
             \overline{\vartheta_{i}(v)}             
$
and
$
S_{2}(u,v) = 
        \sum_{i=1}^{\infty}
             b_{i}
             \vartheta_{i}(u)
             \overline{\vartheta_{i}(v)}  
             ,
$
with $a_{i}, b_{i}\in\mbC$. Then,
\begin{equation}\label{eq_thm_box_prod_spect_1}
\left(
  S_{1}\square S_{2}
\right)(u,v)
   =
   \sum_{i=1}^{\infty}
       a_{i}b_{i}
            \vartheta_{i}(u)
            \overline{\vartheta_{i}(v)}
       .
\end{equation}
Additionally, if
$
S(u,v) = 
        \sum_{i=1}^{\infty}
             \lambda_{i}
                  \vartheta_{i}(u)
                  \overline{\vartheta_{i}(v)}
             ,
$
and $p$ is a polynomial, it follows that
\begin{equation}\label{eq_thm_box_prod_spect_2}
p\left( 
    S^{\square}
\right)
      =
      \sum_{i=1}^{\infty}
             p\left( 
                 \lambda_{i}
              \right)
              \vartheta_{i}(u)
              \overline{\vartheta_{i}(v)}
              .
\end{equation}
\end{theorem}

\begin{proof}
See Appendix~\ref{proof_thm_box_prod_spect}.
\end{proof}


Theorem~\ref{thm_box_prod_spect} is of great value to understand the properties of the box product in terms of spectral decompositions, and while its results are stated for a general symbol $S$, this has immediate consequences when considering $S$ to be a reproducing kernel. In particular,~\eqref{eq_thm_box_prod_spect_2} emphasizes that a polynomial representation transfers from the algebra on the box product, $\ccalA_{K}$, to the spectral domain on a point-wise manner.

We now leverage the algebraic operations stated for $\ccalA_{K}$ for the explicit construction of a homomorphism written in terms of the reproducing property of an RKHS.


\begin{theorem}\label{thm_hom_box_prod}
Let $K: \ccalX\times\ccalX\to\mbC$ be a continuous reproducing kernel inducing the RKHS $\ccalH(K)$, where $\ccalX$ is a compact subset of $\mbR^{d}$. For any polynomial $p(t)\in\mbC[t]$, define $q_{v}\left( K^{\square} \right)=\left(p  \left(K^{\square}\right)  \square K\right)(u,v)$ as the polynomial in the variable $u$ with $v$ held fixed. Then the map $\rho_{K}: \mbC[t]\to\text{End}\left( \ccalH(K)\right)$ defined by
\begin{equation}\label{eq_rho_Ak_Hk}
\rho_{K}
      \left( 
             p\left(
                    t
                \right)
      \right)
               f
             =
             \overline{
             \left\langle 
                     q_{v}\left(  K^{\square} \right)
                    ,
                    f
             \right\rangle}_{\ccalH(K)}
                          =
               g(v)
\end{equation}
is a homomorphism.
\end{theorem}
\begin{proof} 
   See Appendix~\ref{sub_sec_proof_thm_Ak_subsets_in_HK}. 
\end{proof}


\noindent Theorem~\ref{thm_hom_box_prod} emphasizes the role of the homomorphism $\rho_{K}$ as a tool to implement polynomial diffusions in a point-wise manner, and it provides a natural extension of the reproducing property in the RKHS. In fact,~\eqref{eq_rho_Ak_Hk} highlights that the value of a diffused signal at one specific point can be \textit{reproduced} trough the inner product in $\ccalH$ between $q_{v}\left( K^{\square}\right)$ and the signal before the diffusion. This has direct and important implications regarding the computational implementation of diffusions that rely on integral operators, i.e. the computation of diffusions through $\boldsymbol{T}_{K}$ can be efficiently carried out leveraging the RKHS inner product in $\ccalH(K)$. 

One natural questions that opens up from Theorem~\ref{thm_hom_box_prod} is whether an ordinary polynomial diffusion can be computed trough the inner product representation in~\eqref{eq_rho_Ak_Hk}. This will provide a concrete computational tool to compute diffusions -- pointwise -- that involve integral operators. We formalize this fact in the following result.


\begin{theorem}\label{thm_alg_rkhs_vs_alg_classic}
Let $K: \ccalX\times\ccalX\to\mbC$ be a continuous reproducing kernel inducing the RKHS $\ccalH(K)$, and where $\ccalX$ is a compact subset of $\mbR^{d}$. Then, given the ASMs $(\mbC[t], \ccalH(K), \rho_{K})$ and $(\mbC[t], \ccalH(K), \rho)$ with $\rho_{K}$ given by~\eqref{eq_rho_Ak_Hk} and $\rho$ given by $\rho(t)=\boldsymbol{T}_{K}$, we have 
\begin{equation}\label{eq_thm_alg_rkhs_vs_alg_classic}
\rho\left( 
              p\left(
                      t
                \right)      
       \right)
             f
             =
               \rho_{K}
                     \left(
                           p\left( 
                                  t
                              \right)
                     \right)
                           f
                           ,
\end{equation}
for all $f\in\ccalH(K)$ and any polynomial $p(t)\in\mbC[t]$.
\end{theorem}

\begin{proof} See Appendix~\ref{proof_thm_alg_rkhs_vs_alg_classic} \end{proof}



\begin{figure*}
\centering
\definecolor{viridis1}{RGB}{253,231,159}  
\definecolor{viridis2}{RGB}{184,222,166}  
\definecolor{viridis3}{RGB}{123,204,196}  
\definecolor{viridis4}{RGB}{88,171,188}   
\definecolor{viridis5}{RGB}{67,141,165}   
\definecolor{viridis6}{RGB}{72,108,142}   

\begin{tikzpicture}[
	node distance=1.8cm and 1.2cm,
	block/.style={rectangle, draw, minimum width=0.5cm, minimum height=0.5cm, thick, rounded corners=2pt},
	sum/.style={circle, draw, thick, minimum size=0.5cm, fill=white},
	filter/.style={rectangle, draw, minimum width=2.2cm, minimum height=1.4cm, thick, rounded corners=3pt, font=\Large\bfseries},
	arrow/.style={-Stealth, thick},
	label/.style={font=\large}
	]
	
	\node[label] (input) at (0,0) {$f$};
	
	\node[filter, fill=viridis2, below=1cm of input, xshift=1.5cm] (T0) {$\boldsymbol{T}_K^0$};
	\node[block, fill=viridis1, below=1cm of T0] (h0) {$h_0$};
	
	\node[filter, fill=viridis3, below=1cm of input, xshift=5cm] (T1) {$\boldsymbol{T}_K^1$};
	\node[block, fill=viridis2, below=1cm of T1] (h1) {$h_1$};
	
	\node[filter, fill=viridis4, below=1cm of input, xshift=8.5cm] (T2) {$\boldsymbol{T}_K^2$};
	\node[block, fill=viridis3, below=1cm of T2] (h2) {$h_2$};
	
	\node[below=1cm of input, xshift=11cm, font=\LARGE] (dots) {$\cdots$};
	
	\node[filter, fill=viridis6, below=1cm of input, xshift=13.2cm] (TR) {$\boldsymbol{T}_K^R$};
	\node[block, fill=viridis5, below=1cm of TR] (hR) {$h_R$};
	
	\node[sum, below=1cm of hR] (sum) {$+$};
	
	\node[sum, below=1cm of h0] (sum0) {$+$};
	\node[sum, below=1cm of h1] (sum1) {$+$};
	\node[sum, below=1cm of h2] (sum2) {$+$};
	\node[sum, below=3.45cm of dots] (sumdots) {$+$};

	\node[label, right=1.2cm of sum, label=below] (output) {$p(\boldsymbol{T}_K)f$};
	
	\draw[arrow] (input) -| (T0);
	\draw[arrow] (input) -| (T1);
	\draw[arrow] (input) -| (T2);
	\draw[arrow] (input) -| (TR);
	
	\draw[arrow] (T0) -- (h0);
	\draw[arrow] (T1) -- (h1);
	\draw[arrow] (T2) -- (h2);
	\draw[arrow] (TR) -- (hR);
	
	\draw[arrow] (h0) -- (sum0);
	\draw[arrow] (h1) -- (sum1);
	\draw[arrow] (h2) -- (sum2);
	\draw[arrow] (hR) -- (sum);
	
	\draw[arrow] (sum0) -- (sum1);
	\draw[arrow] (sum1) -- (sum2);
	\draw[arrow] (sum2) -- (sumdots);
	\draw[arrow] (sumdots) -- (sum);

	\draw[arrow] (sum) -- (output);
	
	\node[right=0.1cm of T0, font=\small, align=left] at ($(T0)!0.5!(h0)$) {$\mathcal{H}(K)$};
	\node[right=0.1cm of T1, font=\small, align=left]  at ($(T1)!0.5!(h1)$) {$\mathcal{H}(K^{\square 2})$};
	\node[right=0.1cm of T2, font=\small, align=left]  at ($(T2)!0.5!(h2)$){$\mathcal{H}(K^{\square 3})$};
	\node[right=0.1cm of TR, font=\small, align=left]  at ($(TR)!0.5!(hR)$) {$\mathcal{H}(K^{\square (R+1)})$};
	
	\node[left=0.1cm, font=\footnotesize, text=viridis2!70!black] at ($(T0)!0.5!(h0)$) {$\boldsymbol{T}_K^0 f$};
	\node[left=0.1cm, font=\footnotesize, text=viridis3!70!black] at ($(T1)!0.5!(h1)$) {$\boldsymbol{T}_K^1 f$};
	\node[left=0.1cm, font=\footnotesize, text=viridis4!70!black] at ($(T2)!0.5!(h2)$) {$\boldsymbol{T}_K^2 f$};
	\node[left=0.1cm, font=\footnotesize, text=viridis6!70!black] at ($(TR)!0.5!(hR)$) {$\boldsymbol{T}_K^R f$};
	
	\node[left=0.1cm, font=\footnotesize, text=viridis1!70!black] at ($(h0)!0.5!(sum0)$) {$h_0 \boldsymbol{T}_K^0 f$};
    \node[right=0.1cm, font=\footnotesize, text=viridis1!70!black] at ($(h0)!0.5!(sum0)$) {$\displaystyle h_0 \sum_{\ell_{0}}\alpha_{\ell_{0}}k_{\ell_{0}}$};
	\node[left=0.1cm, font=\footnotesize, text=viridis2!70!black] at ($(h1)!0.5!(sum1)$) {$h_1 \boldsymbol{T}_K^1 f$};
    \node[right=0.1cm, font=\footnotesize, text=viridis2!70!black] at ($(h1)!0.5!(sum1)$) {$\displaystyle h_1 \sum_{\ell_{1}}\alpha_{\ell_{1}}k_{\ell_{1}}^{(2)}$};
	\node[left=0.1cm, font=\footnotesize, text=viridis3!70!black] at ($(h2)!0.5!(sum2)$) {$h_2 \boldsymbol{T}_K^2 f$};
    \node[right=0.1cm, font=\footnotesize, text=viridis3!70!black] at ($(h2)!0.5!(sum2)$) {$\displaystyle h_2 \sum_{\ell_{2}}\alpha_{\ell_{2}}k_{\ell_{2}}^{(3)}$};
	\node[left=0.1cm, font=\footnotesize, text=viridis5!70!black] at ($(hR)!0.5!(sum)$) {$h_R \boldsymbol{T}_K^R f$};
    \node[right=0.1cm, font=\footnotesize, text=viridis5!70!black] at ($(hR)!0.5!(sum)$) {$\displaystyle h_R \sum_{\ell_{R}}\alpha_{\ell_{R}}k_{\ell_{R}}^{(R+1)}$};
	
	
\end{tikzpicture} 
    \caption{Illustration diagram of the results in Theorem~\ref{thm_TKn_vs_Kboxn} and Corollary~\ref{cor_diff_as_sum_Hkn}. The operator $\boldsymbol{T}^{r}_{K}$ maps the signal $f\in\ccalH(K)$ on the RKHS $\ccalH\left(K^{\square (r+1)}\right)$, where the resultant diffused signal can be written as an expansion in terms of the kernel functions $k_{\ell_{r}}^{(r+1)}$. Then, the result of diffusing $f$ with $p(\boldsymbol{T}_{K})=\sum_{r=0}^{R}h_{r}\boldsymbol{T}_{K}^{r}f$ is obtained as a weighted sum of signals on the RKHS spaces $\{ \ccalH(K^{\square (r+1)}) \}_{r=0}^{R}$, where the weights are the filter coefficients $\{ h_{r} \}_{r=0}^{R}$.}
    \label{fig_rkhs_bank_filter}
\end{figure*}
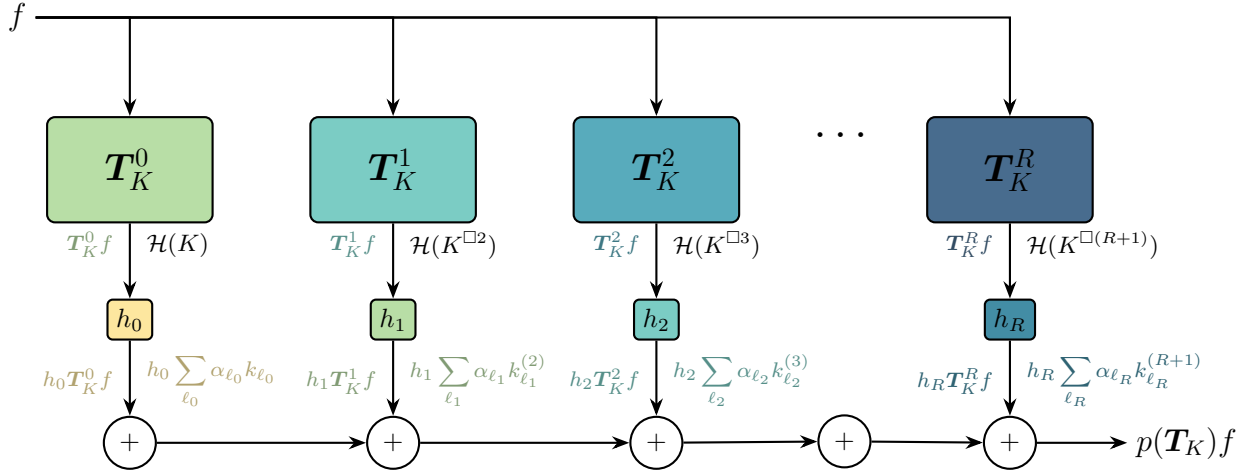



\subsubsection{\underline{Implications in Gphon-SP}}

In Theorem~\ref{thm_K_from_W_gphonsp} we determined that any graphon $W$ induces the reproducing kernel $K=W\square W$. However, using Theorems~\ref{thm_alg_RKHS_box_prod_1} and~\ref{thm_box_prod_spect} we can extend the connection between a graphon and a myriad of reproducing kernels. This idea is formalized as follows.


\begin{corollary}\label{corll_K_from_Wboxn}
Let $W(u,v)$ be a graphon with eigenfunctions $\{ \varphi_{i}(u)\}_{i=1}^{\infty}$ and eigenvalues $\{ \lambda_i \}_{i=1}^{\infty}$. For any integer $n\geq 1$, the function $K:[0,1]\times [0,1]\to [0,1]$ given by
\begin{equation}\label{eq_corll_K_from_Wboxn_1}
K(u,v) =
        W^{\square 2n}
       =
        \sum_{i=1}^{\infty}
             \lambda_{i}^{2n}
             \varphi_{i}(u)
             \varphi_{i}(v)
             ,
\end{equation}
where $W^{\square 2n}$ denotes the $2n$-fold box product of $W$ with itself, is a reproducing kernel. Moreover, the associated integral operator satisfies
\begin{equation}\label{eq_corll_K_from_Wboxn_2}
\boldsymbol{T}_{K}
     =
     \boldsymbol{T}_{W}^{2n}
     .
\end{equation}
\end{corollary}

\begin{proof}
    This follows directly from the application of Theorem~\ref{thm_box_prod_spect} jointly with $K=W\square W$.
\end{proof}


\noindent This result follows directly from the fact that $W^{\square 2}$ is a reproducing kernel, and $W^{\square 2n}$ is obtained by taking the $n$-fold box product of $W^{\square 2}$ with itself, which preserves the reproducing kernel property. Equation~\eqref{eq_corll_K_from_Wboxn_2} follows immediately from~\eqref{eq_thm_K_from_W_gphonsp_2} in Theorem~\ref{thm_K_from_W_gphonsp}. A key insight from Corollary~\ref{corll_K_from_Wboxn} is that functions on the graphon space admit representations in terms of multiple reproducing kernels. That is, the graphon serves as a generating function for a family of RKHS spaces whose elements can be combined to represent graphon signals---a point we generalize in the next section.

Given the natural connection between a graphon $W$ and its induced reproducing kernel $K=W\square W$, Theorem~\ref{thm_alg_rkhs_vs_alg_classic} has immediate applications in Gphon-SP. More specifically, even when a graphon representation is chosen to model a large graph, the computation of diffusions through the graphon shift operator can be efficiently carried out leveraging the RKHS inner product associated to the reproducing kernel $K$.

On the other hand, an aspect of great value from Theorem~\ref{thm_alg_rkhs_vs_alg_classic} lies on the complete equivalence between the classical diffusions and the point-wise diffusions associated with the RKHS. Additionally, in the light of Theorem~\ref{thm_alg_rkhs_vs_alg_classic} one can consider that the RKHS $\ccalH(K)$ is an invariant subspace to the action of the diffusions given by $\rho\left(p\left(t\right)\right)f$ in $L_{2}(\ccalT,\mu)$.




\section{RKHS filter representation and Design of Algebraic Convolutional Filters}
\label{sec_rkhs_filt_rep_asp_filt_desgn}


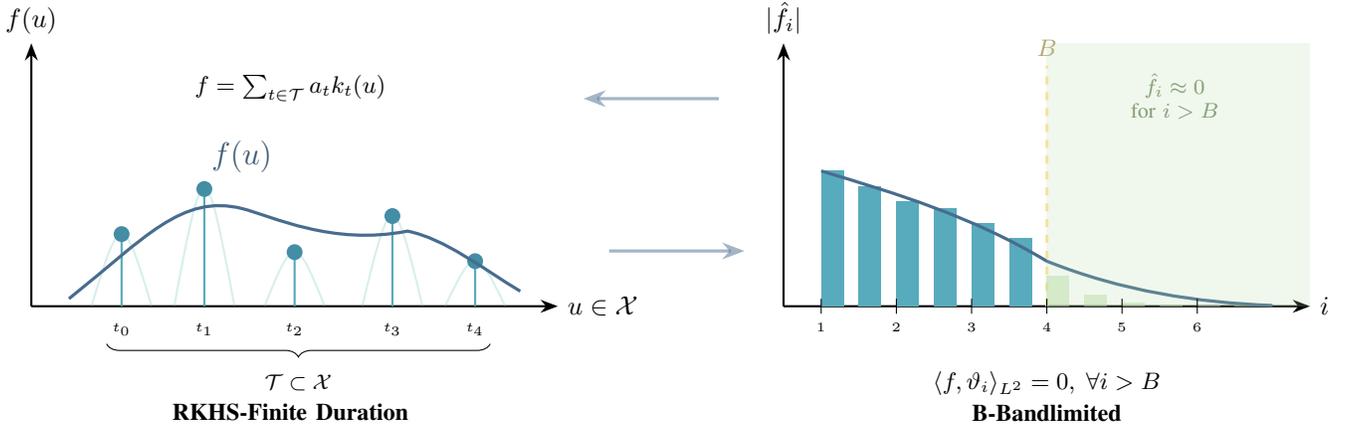
\begin{figure*}
\centering
\definecolor{viridis1}{RGB}{253,231,159}  
\definecolor{viridis2}{RGB}{184,222,166}  
\definecolor{viridis3}{RGB}{123,204,196}  
\definecolor{viridis4}{RGB}{88,171,188}   
\definecolor{viridis5}{RGB}{67,141,165}   
\definecolor{viridis6}{RGB}{72,108,142}   

\begin{tikzpicture}[
	>=Stealth,
	axis/.style={thick,->,>=Stealth},
	stem/.style={thick},
	label/.style={font=\small}
	]
	
	\begin{scope}[local bounding box=spatial]
		\draw[axis] (0,0) -- (7,0) node[right] {$u \in \mathcal{X}$};
		\draw[axis] (0,0) -- (0,3.5) node[above] {$f(u)$};
		
		\def\supportpoints{{1.2, 2.3, 3.5, 4.8, 5.9}}
		\def\coefficients{{0.8, 1.3, 0.6, 1.0, 0.5}}
		
		\foreach \i/\pos/\coef in {0/1.2/0.8, 1/2.3/1.3, 2/3.5/0.6, 3/4.8/1.0, 4/5.9/0.5} {
			\pgfmathsetmacro{\height}{\coef*1.2}
			\draw[viridis3, thick, opacity=0.3] 
			(\pos-0.4,0) .. controls (\pos-0.2,\height*0.7) and (\pos-0.1,\height) .. 
			(\pos,\height) .. controls (\pos+0.1,\height) and (\pos+0.2,\height*0.7) .. 
			(\pos+0.4,0);
		}
		
		\foreach \i/\pos/\coef in {0/1.2/0.8, 1/2.3/1.3, 2/3.5/0.6, 3/4.8/1.0, 4/5.9/0.5} {
			\pgfmathsetmacro{\height}{\coef*1.2}
			\draw[stem, viridis4] (\pos,0) -- (\pos,\height);
			\fill[viridis5] (\pos,\height) circle (3pt);
			\node[below, font=\tiny] at (\pos,-0.1) {$t_{\i}$};
		}
		
		\draw[viridis6, very thick, smooth] 
		(0.5,0.1) .. controls (1.5,0.9) and (2.0,1.5) .. 
		(2.8,1.3) .. controls (3.2,1.2) and (4.0,0.8) .. 
		(5.0,1.0) .. controls (5.5,0.9) and (6.0,0.5) .. (6.5,0.2);
		
		\node[viridis6!80!black, font=\large\bfseries] at (2.8,2.0) {$f(u)$};
		
		\node[font=\small, align=center] at (3.5, 2.9) {
			$f = \sum_{t \in \mathcal{T}} a_t k_t(u)$
		};
		
		\node[label, align=center] at (3.5, -1.4) {
			\textbf{RKHS-Finite Duration}
		};
		
		\draw[decorate,decoration={brace,amplitude=5pt,mirror}] 
		(1.0,-0.5) -- (6.1,-0.5) node[midway,below=8pt,font=\footnotesize] {$\mathcal{T} \subset \mathcal{X}$};
	\end{scope}
	
	\begin{scope}[xshift=10cm, local bounding box=spectral]
		\draw[axis] (0,0) -- (7,0) node[right] {$i$};
		\draw[axis] (0,0) -- (0,3.5) node[above] {$|\hat{f}_i|$};
		
		\def\bandwidth{3.5}
		\draw[viridis1, very thick, dashed] (\bandwidth,0) -- (\bandwidth,3.2);
		\node[above, viridis1!70!black, font=\small\bfseries] at (\bandwidth,3.2) {$B$};
		
		\foreach \i/\height in {0.5/1.8, 1.0/1.6, 1.5/1.4, 2.0/1.3, 2.5/1.1, 3.0/0.9, 3.5/0.4, 4.0/0.15, 4.5/0.05, 5.0/0.02, 5.5/0.01, 6.0/0.005} {
			\pgfmathparse{\i < \bandwidth ? 1 : 0}
			\ifnum\pgfmathresult=1
			\fill[viridis4] (\i,0) rectangle (\i+0.3,\height);
			\else
			\fill[viridis2, opacity=0.5] (\i,0) rectangle (\i+0.3,\height);
			\fi
		}
		
		\draw[viridis6, very thick, smooth] 
		(0.5,1.8) .. controls (1.5,1.5) and (2.5,1.2) .. 
		(3.5,0.6) .. controls (4.5,0.2) and (5.5,0.05) .. (6.5,0.01);
		
		\fill[viridis2, opacity=0.2] (\bandwidth,0) rectangle (7,3.5);
		\node[viridis2!70!black, font=\footnotesize, align=center] at (5.2,2.8) {
			$\hat{f}_i \approx 0$\\
			for $i > B$
		};
		
		\node[label, align=center] at (3.5, -1.2) {
			$\langle f, \vartheta_i \rangle_{L^2} = 0, \; \forall i > B$\\[2pt]
			\textbf{B-Bandlimited}
		};
		
		\foreach \i in {1,...,6} {
			\draw (\i-0.5,-0.1) -- (\i-0.5,0.1);
			\node[below, font=\tiny] at (\i-0.5,-0.1) {$\i$};
		}
	\end{scope}
	
	\coordinate (leftcenter) at ($(spatial.east)+(0.5,1.5)$);
	\coordinate (rightcenter) at ($(spectral.west)+(-0.5,1.5)$);
	
	\draw[->,very thick, viridis6,opacity=0.5] (leftcenter)++(-1,-2) -- ++(1.8,0) 
	node[midway,above,sloped,font=\small\bfseries] {};
	\draw[->,very thick, viridis6,opacity=0.5] (rightcenter) -- ++(-1.8,0);
	

\end{tikzpicture} 
    \caption{Diagramatic illustration of the consequences of Corollary~\ref{cor_rkhs_uncertainty}. An RKHS with a finite expansion in terms of the $k_{t}$ functions cannot be exactly bandlimited. However, due to the nature of the eigenvalues, $\sigma_i$, of the operator $\boldsymbol{T}_{K}$ the signal can be \textit{approximately} bandlimited. Given the decomposition of a signal $f\in\ccalH(K)$ with the representation in~\eqref{eq_f_decomp_mult_bwd}, the tradeoff between RKHS finiteness and bandlimited frequencies is determined by $\vert\ccalT\vert-B$. If $B<\vert\ccalT\vert$ it is possible to choose $\vert\ccalT\vert -B$ coefficients $a_{t}$ to minimize the size of $f_{B+1 : \vert\ccalT\vert}$ in~\eqref{eq_f_decomp_mult_bwd}. Then, $B$ coefficients determine a specific low pass behavior, while the remaining coefficients are selected to reduce the size of the residuals (green shaded area on the right side). This implies that, the larger $\vert\ccalT\vert - B$ is, the faster the residuals can go to zero.}
    \label{fig_bandlimited_tradeoff}
\end{figure*}


The direct connection between the box product and spectral representations on an RKHS naturally extends to the algebraic structure of convolutional filters when expressed as polynomials of $\boldsymbol{T}_{K}$. While this fact follows immediately from the results stated in the previous sections, it is worth emphasizing that the box product is closed with respect to the production of reproducing kernels---that is, the box product of a reproducing kernel with itself yields another reproducing kernel. This property, together with the spectral representation associated with the reproducing kernel, allows us to establish the connection between the diffusion processes induced by the different kernels generated through the box product. We formalize these ideas in the following result.


\begin{theorem}\label{thm_TKn_vs_Kboxn}
Let $K$ be a continuous kernel on $\ccalX\times\ccalX$, $\mu$ be a finite Borel measure on the compact set $\ccalX\subset\mbR^{d}$. Then, it follows that
\begin{equation}
\boldsymbol{T}_{K}^{n}
     f
      =
       \boldsymbol{T}_{K^{\square n}}
       f
       .
\end{equation}
Additionally, if $p(t)\in\mbC[t]$, then it follows that
\begin{equation}
p\left(
         \boldsymbol{T}_{K}
  \right)f
        =
         \boldsymbol{T}_{p\left(K^{\square r}\right)}
                                f   
          .
\end{equation}
\end{theorem}

\begin{proof}
See Appendix~\ref{proof_thm_TKn_vs_Kboxn}.
\end{proof}


\noindent While the results in Theorem~\ref{thm_TKn_vs_Kboxn} are interesting in their own right, their true value lies in enabling the interpretation of filtered signals as decompositions into sums of RKHS subspaces of $L_{2}(\ccalX,\mu)$. This interpretation follows from two key facts: first, $K^{\square n}$ is the reproducing kernel of an RKHS $\ccalH\left(K^{\square n}\right)\subset L_{2}(\ccalX,\mu)$; second, $\ccalH\left(K^{\square n}\right)$ coincides with the range of the operator $\boldsymbol{T}_{K^{\square (n-1)}}: L_{2}(\ccalX,\mu)\to L_{2}(\ccalX,\mu)$~\cite{paulsen2016introduction,wainwright2019high,ghojogh2021reproducing}. We formalize this observation in the following corollary.


\begin{corollary}\label{cor_diff_as_sum_Hkn}
Let $K: \ccalX\times\ccalX\to\mbC$ be a continuous reproducing kernel inducing the RKHS $\ccalH(K)$, and where $\ccalX$ is a compact subset of $\mbR^{d}$. If $f\in\ccalH(K)$, then the signal $g=\sum_{r=0}^{R}h_{r}\boldsymbol{T}_{K}^{r}f$ can be written as an element of the sum of vector spaces
\begin{equation}\label{eq_cor_diff_as_sum_Hkn_1}
\ccalH = \ccalH(K)+\ccalH\left(K^{\square 2}\right)
         +\ldots+
         \ccalH\left(K^{\square R}\right)
         +
         \ccalH\left(K^{\square (R+1)}\right)
         .
\end{equation}
Additionally, $g\in\ccalH$ can be written as
\begin{equation}\label{eq_cor_diff_as_sum_Hkn_2}
g(u) =
   \sum_{r=0}^{R}
        h_{r}
        \sum_{\ell_{r}}
        \alpha_{\ell_{r}}
         k_{\ell_{r}}^{(r+1)}(u)
         ,
\end{equation}
where $k_{\ell_{r}}^{(r+1)}(u)=K^{\square (r+1)}\left(u,\ell_{r}\right)$ and $\alpha_{\ell_{r}}\in\mbR$.
\end{corollary}

\begin{proof}
The proofs follows directly from the application of Theorem~\ref{thm_TKn_vs_Kboxn}.
\end{proof}


\noindent From~\eqref{eq_cor_diff_as_sum_Hkn_2} we obtain a more explicit implementation of polynomial diffusions as point-wise representations. In fact,~\eqref{eq_cor_diff_as_sum_Hkn_2} comes from a direct application of the reproducing property in the RKHS spaces involved in the sum~\eqref{eq_cor_diff_as_sum_Hkn_1}. This provides a way for an efficient and compact implementation of any polynomial diffusion when the reproducing kernels
$K^{\square n}$ are computed before hand. Fig.~\ref{fig_rkhs_bank_filter} lays out a natural bank filter interpretation of~\eqref{eq_cor_diff_as_sum_Hkn_2}.


\subsection{Spatial-Spectral Localization}
\label{sub_sec_spat_spec_loclz}

The spectral representation in~\eqref{eq_thm_box_prod_spect_2} and the point-wise implementation of the filters in~\eqref{eq_rho_Ak_Hk} allows us to establish a direct relationship between the effect of the filter in the spectral domain with respect to its expansion in terms of the $k_{v}$ functions. We formalize such relationship in the following corollary.


\begin{corollary}\label{cor_spec_vs_ku}
Let $p(t)$ be a polynomial and $q_{v}\left( K^{\square} \right)=\left(p  \left(K^{\square}\right)  \square K\right)(u,v)$ be given according to Theorem~\ref{thm_hom_box_prod}. If $q_{v}\left( K^{\square}\right)=\sum_{\ell\in\ccalL}\alpha_{\ell}k_{\ell}$, then it follows that
\begin{equation}\label{eq_lemma_spec_vs_ku}
p\left( 
     \sigma_{i}
 \right)
        =
         \frac{1}{\overline{\vartheta_{i}(v)}}
         \sum_{\ell\in\ccalL}
              \alpha_{\ell}
              \overline{\vartheta_{i}(\ell)}
              ,
\end{equation}
for all $\vartheta_{i}(v)\neq 0$.
\end{corollary}

\begin{proof} 
The proof follows from Theorem~\ref{thm_box_prod_spect} considering the polynomial representation of $q_{v}\left( K^{\square}\right)$ with a continuous kernel $K$.
\end{proof}


\noindent This result establishes a characterization of the spectral response of a filter at particular frequencies in terms of the coefficients that determine the expansion of $q_{v}$ in terms of the kernel functions. More specifically, the right hand side of~\eqref{eq_lemma_spec_vs_ku} states that the response of the filter at $\sigma_{i}$ is determined by a discrete inner product between $\{ \alpha_\ell \}_{\ell\in\ccalL}$ and the samples of $\vartheta_{i}$ at $\ccalL$, $\{ \vartheta_{i}(\ell) \}_{\ell\in\ccalL}$. Then, the amplitude of the filter at a given frequency is given by the projection of $\{ \alpha_\ell \}_{\ell\in\ccalL}$ on $\{ \vartheta_{i}(\ell) \}_{\ell\in\ccalL}$. This entails an interpretation in terms of similarity, i.e. the behavior of $\sum_{\ell\in\ccalL}\alpha_{\ell}k_{\ell}$ will be determined by the correlation between the sequences $\{ \alpha_\ell \}_{\ell\in\ccalL}$ and $\{ \vartheta_{i}(\ell) \}_{\ell\in\ccalL}$. A high correlation leads to a large value of the filter at $\sigma_{i}$, while if $\{ \alpha_\ell \}_{\ell\in\ccalL}$ and $\{ \vartheta_{i}(\ell) \}_{\ell\in\ccalL}$ are orthogonal (unrelated) the value $p(\sigma_{i})$ will be small. These insights collectively enable a clear interpretation of $\{ \alpha_\ell \}_{\ell\in\ccalL}$ in both the frequency domain and the original domain of the signals.

With the results we have at hand, we now proceed to establish the tradeoffs between the localization on frequency and the original domain. To this end, we introduce some basic terminology. 


\begin{definition}
    Let $K: \ccalX\times\ccalX\to\mbC$ be a continuous reproducing kernel inducing the RKHS $\ccalH(K)$, and where $\ccalX$ is a compact subset of $\mbR^{d}$. If $\vartheta_{i}$ is the $i$-th eigenvector of the integral operator $\boldsymbol{T}_{K}$, then we say that $f\in\ccalH(K)$ is $B$-bandlimited in frequency if 
\begin{equation}
\widehat{f}_{i}
         =
           \left\langle
                 f
                 ,
                 \vartheta_{i}
           \right\rangle_{L_{2}}
                           =
                           0,
                           \quad
                           \forall i>B
                           ,
\end{equation}
where the indices ``$i$" of eigenvectors are determined by the ordering implied from $ \sigma_{1} \geq \sigma_{2} \geq \cdots$. Additionally, we say that $f\in\ccalH(K)$ is of RKHS-finite duration if there exists a finite set $\ccalT\subset\ccalX$ such that
\begin{equation}
f =
    \sum_{t\in\ccalT}
           a_{t}
           k_{t}
           ,
\end{equation}
where $a_{t}\in\mbC$.
\end{definition}


\noindent The notion of a signal being bandlimited, establishes a finite degree of freedom on the frequency domain, while the notion of finite RKHS-duration implies a finite degree of freedom when represented in terms of the functions $k_{v}$. In almost all signal models, it is convenient to have signals that are bandlimited, not only for computation purposes, but to establish limits of error when understading properties like stability and transferability~\cite{gphon_pooling_j,gphon_samp,algnn_nc_j,parada_algnn,msp_j}. On the other hand, one of the more appealing uses of the RKHS theory is precisely having a finite expansion -- in terms of the $k_{v}$ -- of any quanity of interest. In the following result we establish the conditions on a signal to have both attributes.


\begin{corollary}\label{cor_rkhs_uncertainty}
Let $K$ be a continuous kernel on $\ccalX\times\ccalX$, $\mu$ be a finite Borel measure on the compact set $\ccalX\subset\mbR^{d}$. Let $f = \sum_{t\in\ccalT}a_{t}k_{t}$, where $\ccalT\subset\ccalX$ is finite. If $f$ is $B$-bandlimited in frequency, it follows that
\begin{equation}\label{eq_proof_thm_rkhs_uncertainty_4}
              \sigma_{i}
              \sum_{t\in\ccalT}
                        a_{t}
                        \overline{\vartheta_{i}(t)}
                        =
                        0
                        \quad\forall i>B
                        .
\end{equation}
\end{corollary}

\begin{proof}
See Appendix~\ref{subsec_cor_rkhs_uncertainty}.
\end{proof}


\noindent From~\eqref{eq_proof_thm_rkhs_uncertainty_4} we can see that when $\sigma_{i}\neq 0$ for all finite $i$, it is not possible to have a bandlimited signal that is simultaneously RKHS-finite. However, the compactness of the operator $\boldsymbol{T}_{K}$ ensures that $0$ is an accumulation point of the $\{ \sigma_{i} \}_{i=1}^{\infty}$, and therefore the magnitude of $\sigma_{i}$ decreases as the index $i$ increases. This particular attribute guarantees that $\sigma_{i}\sum_{t\in\ccalT}a_{t}\overline{\vartheta_{i}(t)}$ decreases its value when $i$ increases, which highlights the fact that it is possible to have a finite -- RKHS signal that is \textit{approximately} bandlimited.

Then, when considering approximately bandlimited signals that are also RKHS finite, there are some fundamental tradeoffs between the size of $\ccalT$ and the frequency bandwidth $B$. To show this, let us start pointing out that if $f$ is as considered in Corollary~\ref{cor_rkhs_uncertainty}, we can understand $f$ in terms of the decomposition $f = f_{1 : B} + f_{B+1 : \vert\ccalT\vert} + f_{\vert\ccalT\vert+1 : \infty},$ where
\begin{equation}\label{eq_f_decomp_mult_bwd}
f_{L_i : L_s}
   =
    \sum_{i=L_i}^{L_s}
           \left( 
               \sigma_{i}
               \sum_{t\in\ccalT}
                     a_{t}
                     \overline{\vartheta_{i}(t)}
           \right)
           .
\end{equation}
The term $f_{1:B}$ describes the low pass behavior of the signal $f$, while the terms $f_{B+1 : \vert\ccalT\vert}$ and $f_{\vert\ccalT\vert+1 : \infty}$ are residuals. If $B<\vert\ccalT\vert$ it is possible to choose $\vert\ccalT\vert -B$ coefficients $a_{t}$ to minimize the size of $f_{B+1 : \vert\ccalT\vert}$, i.e. $B$ coefficients determine a specific low pass behavior, while the remaining coefficients are selected to reduce the size of the residuals. This implies that, the larger $\vert\ccalT\vert - B$ is, the faster the residuals can go to zero. Additionally, notice that while choosing the $a_{t}$ is crucial to reduce the size of $f_{B+1 : \vert\ccalT\vert}$, we can afford an arbitrary choice of $a_{t}$ regarding $f_{\vert\ccalT\vert+1 : \infty}$ as a consequence of the decay rate of $\sigma_i$. This Analysis prompts a fundamental fact about the bandwidth of an approximately bandlimited signal in an RKHS, its bandwidth is upper bounded by the number of terms used in an expansion on the $k_v$ functions.


\subsection{Spectral Filters as Learnable Functions in the RKHS}


\begin{figure*}
%
%
    \centering
       \begin{subfigure}{.32\linewidth}
	   \centering
          \includegraphics[width=1\textwidth]{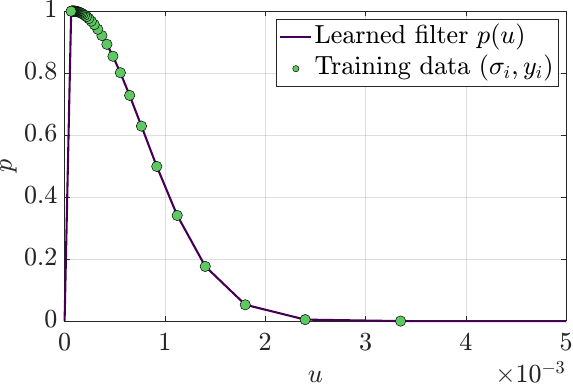} 
          \caption*{$q=35$}
       \end{subfigure}
    	\begin{subfigure}{.32\linewidth}
		\centering
            \includegraphics[width=1\textwidth]{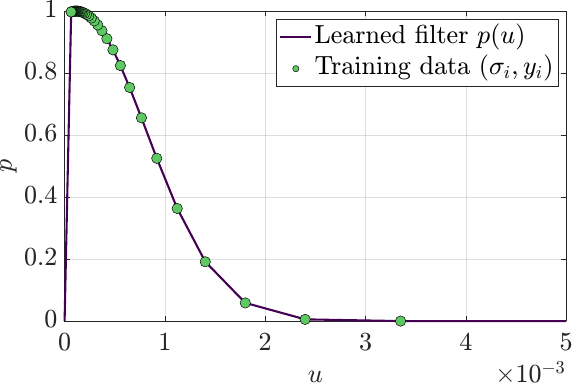}
            \caption*{$q=30$}
\end{subfigure}
    	\begin{subfigure}{.32\linewidth}
		\centering
  \includegraphics[width=1\textwidth]{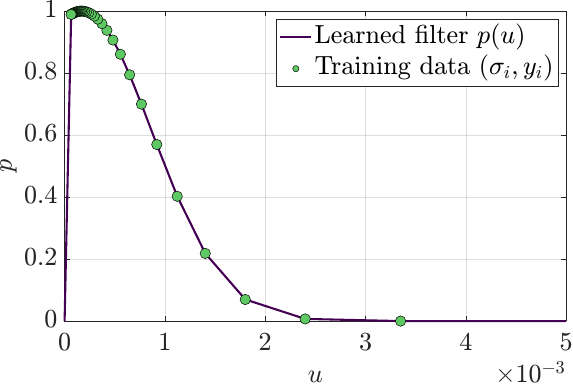} 
  \caption*{$q=25$}
\end{subfigure}
%
%
\centering
\begin{subfigure}{.32\linewidth}
        \centering
        \includegraphics[width=1\textwidth]{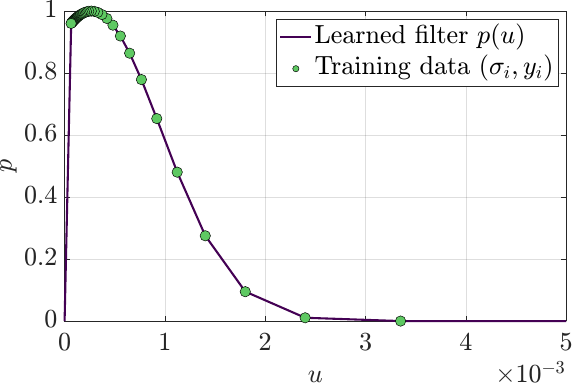} 
        \caption*{$q=20$}
\end{subfigure}
\begin{subfigure}{.32\linewidth}
		\centering
            \includegraphics[width=1\textwidth]{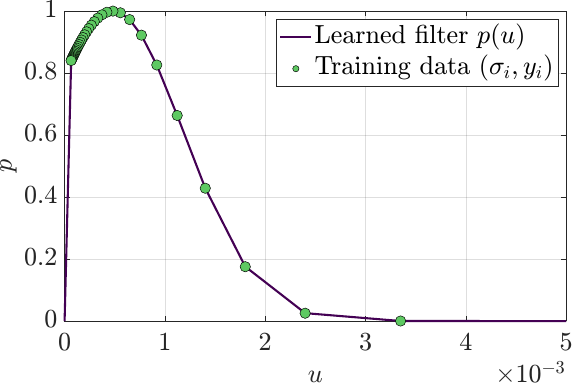} 
            \caption*{$q=15$}
\end{subfigure}
\begin{subfigure}{.32\linewidth}
		\centering
            \includegraphics[width=1\textwidth]{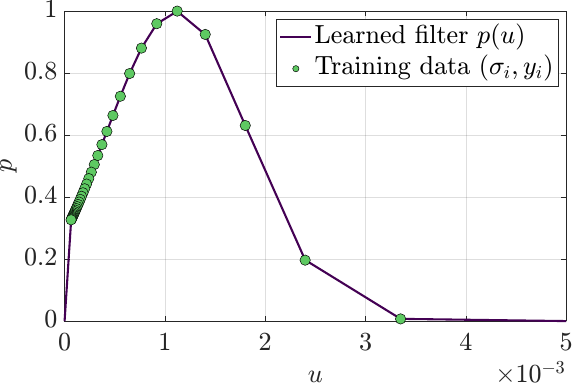} 
            \caption*{$q=32$}
\end{subfigure}
    \caption{Illustration associated to Example~\ref{exmpl_rkhs_reg_filter}. Using the representer theorem we find the optimal polynomial filter $p(t)$, such that $p(\sigma_{i})=\exp(-(\sigma_{i}-\sigma_{c})^2/\gamma)$ where $\sigma_c$ and $\gamma$ are fixed parameters. From Corollary~\ref{cor_filt_representer}, the optimal filter can be written as $p^{\ast}(u)=\sum_{i=1}^{q}a_{i}k_{\sigma_{i}}(u)$. The pictures show the resultant filter, when considering multiple values of $q$ -- the number of points used in the regression -- and multiple values of $\sigma_{c}$ (the localization of the maximum point of the filters).  }
    \label{fig_rkhs_filt_regression}
\end{figure*}


Given concrete data associated with a specific ASM, one can design a filter with the aim of obtaining a particular result when filtering a signal. This is specially important when ASM are used to define convolutional neural networks, where the learnable parameters are the filters in the ASM~\cite{parada_algnn, algnn_nc_j}. In the ASM $\left(\mbC[t], \ccalH(K), \rho \right)$ with $\rho(t)=\boldsymbol{T}_{K}$, one can understand the effect of a filter in the spectral domain by means of
\begin{equation}
p\left( 
      \boldsymbol{T}_{K}
  \right)
      f
      =
      \sum_{i=1}^{\infty}
             p\left( 
                  \sigma_{i}
               \right)    
                  \left\langle
                        f
                        ,
                        \vartheta_i
                  \right\rangle_{L_{2}}
                  \vartheta_{i}
                  , 
\end{equation}
where it is clear that the tangible effect of a filter on a signal is subordinated to the evaluation of the filter on the spectrum of $\boldsymbol{T}_{K}$, $\{ \sigma_i \}_{i=1}^{\infty}$. This emphasizes that when designing and/or learning a filter with specific features, one can focus the effort on minimizing a point-wise error function on $\{ \sigma_i \}_{i=1}^{\infty}$. In the following result, we show that for specific choices of the domain of the signals, the designed/learned filter can be expressed as a decomposition in terms of the functions $k_{v}$ centered on $\{ \sigma_i \}_{i=1}^{\infty}$.


\begin{corollary}\label{cor_filt_representer}
Let $K: \ccalX\times\ccalX\to\mbR$ be a continuous reproducing kernel inducing the RKHS $\ccalH(K)$, where $\ccalX\subset\mbR^{+}$ is compact. Let $\{ \sigma_i \}_{i=1}^{\infty}\subset\ccalX$ be the eigenvalues of $\boldsymbol{T}_{K}: L^{2}(\ccalT,\mu)\to L^{2}(\ccalT,\mu)$, and let $\{ y_{i} \}_{i=1}^{q}\subset\mbR$ be the desired values of a filter on $\{ \sigma_i \}_{i=1}^{q}$. Given $g: [0,\infty]\to\mbR$ a strictly increasing real valued function, and $E:\left(\ccalX, \mbR^2 \right)^{q} \to \mbR\cup\infty $ an arbitrary error function, if
\begin{multline}\label{eq_or_filt_representer_1}
p^{\ast}
    =
     \argmin_{p\in\ccalH(K)}
             \left\lbrace 
                   E\left( 
                          \left(\sigma_{1},y_{1},p(\sigma_1) \right)
                          ,
                          \ldots
                          ,
                          \left(\sigma_{q},y_{q},p(\sigma_q) \right)
                    \right)
                    \right.
                    \\
                    +
                    \left.
                    g\left( 
                         \Vert 
                             p
                         \Vert_{\ccalH}
                     \right)
             \right\rbrace
             ,
\end{multline}
is the optimal filter in $\ccalH(K)$ that minimizes the regularized error function in~\eqref{eq_or_filt_representer_1}, then
\begin{equation}\label{eq_or_filt_representer_2}
p^{\ast}(u)
     =
      \sum_{i=1}^{q}
            a_{i}
            k_{\sigma_{i}}(u)
            ,
\end{equation}
where $a_{i}\in\mbR$ for all $i=1,\ldots,q$.
\end{corollary}

\begin{proof} 
The proof follows directly from the application of the representer theorem, see~\cite{paulsen2016introduction}.
\end{proof}


\noindent Notice that the result in Corollary~\ref{cor_filt_representer} follows from the application of the representer theorem (RT) and the fact that $\{ \sigma_{i} \}_{i=1}^{\infty}\subset\ccalX$. The term $g\left(\Vert p\Vert_{\ccalH}\right)$ in~\eqref{eq_or_filt_representer_1} plays the role of a regularizer when finding the optimal $p^{\ast}\in\ccalH(K)$ that minimizes the error. Classical choices of $g$ and $E$ include $g(u)=\lambda u^{2} $ and 
\begin{equation}
E\left( 
      \left(\sigma_{1},y_{1},f(\sigma_1) \right)
      ,
      \ldots
      ,
      \left(\sigma_{q},y_{q},f(\sigma_q) \right)
  \right)  
        =
         \sum_{i=1}^{q}
                \left( 
                    y_{i}
                    -
                    f(\sigma_i)
                \right)^{2}
                .
\end{equation}

While the use of the RT does not provide itself anything new regarding RKHS, its particular application in Corollary~\ref{cor_filt_representer} emphasizes that the filters in the ASM can be learned/designed as elements of the RKHS, i.e. an element of $\mbC[t]$ is approximated by an element of $\ccalH(K)$. This constitutes an advantage from a computational point of view, and has direct implications in neural networks. In fact, this implies that -- satisfied the conditions of Corollary~\ref{cor_filt_representer} -- the spectral representation of the filters is has a finite dimensional representation on the original domain $\ccalX$. It is important to emphasize that this result has profound implications in Gphon-SP, in those scenarios where the graphon is itself a reproducing kernel. For instance, when learning the parameters of a Gphon neural network one can reformulate the learning of the polynomial filters as a classical RKHS learning problem. The following example illustrates a numerical experiment performed to learn the filters following Corollary~\ref{cor_filt_representer}.


\begin{example}\normalfont 
\label{exmpl_rkhs_reg_filter}
Let us consider the graphon and kernel $K=W=\min (u,v)$, whose eigenvalues and eigenvectors are presented in Example~\ref{exm_cor_gphon_fourier_kv}. We rely on the representer theorem to find the optimal polynomial filter $p(t)$, such that the amplitudes on the eigenvalues $\sigma_{i}$, are tied to $p(\sigma_{i})=\exp(-(\sigma_{i}-\sigma_{c})^2/\gamma)$ where $\sigma_c$ and $\gamma$ are fixed parameters. As per Corollary~\ref{cor_filt_representer}, such optimal filter can be written as $p^{\ast}(u)=\sum_{i=1}^{q}a_{i}k_{\sigma_{i}}(u)$. Figure~\ref{fig_rkhs_filt_regression} depicts the resultant filter, when considering multiple values of $q$ -- the number of points used in the regression -- and multiple values of $\sigma_{c}$ which lead to different filter types.  
\end{example}









\section{Discussion and Conclusions}
\label{sec_discussion}

This paper establishes a comprehensive theoretical framework connecting integral operators, reproducing kernel Hilbert spaces, and algebraic signal processing. Our analysis reveals that integral-operator-based filtering naturally induces RKHS structures through the box product operation, providing an alternative yet equivalent perspective to classical operator-based implementations. The results complement our prior work~\cite{rkhs_conv}, which demonstrated how RKHS with domain-based algebraic structure induce convolutional algebras. In contrast, this paper shows that integral operators themselves carry intrinsic algebraic structure—embodied in the box product algebra—that enables polynomial filtering, spectral analysis, and learning directly within RKHS.

The central insight is that the range of an integral operator $\boldsymbol{T}_S$ with symbol $S(u,v)$ naturally forms an RKHS with reproducing kernel $K = S \Box S^*$. In graphon signal processing, this perspective reveals that each polynomial diffusion $p(\boldsymbol{T}_W)$ produces signals in a specific RKHS with kernel $K = W\square W$, establishing precise relationships between graphon eigenspaces and RKHS representations that extend naturally to directed graphons.

A key contribution is the characterization of the box product algebra $\mathcal{A}_K$ induced by reproducing kernels. We proved that iterated box products $K^{\Box r}$ generate a unital algebra where polynomial operations correspond to polynomial filtering (Theorem~\ref{thm_alg_RKHS_box_prod_1}), and that polynomials in the box product admit eigenfunction expansions with coefficients given by polynomials of the associated eigenvalues (Theorem~\ref{thm_box_prod_spect}). This algebraic structure arises intrinsically from integral operators rather than from operations on the signal domain, providing a natural framework for understanding how filtering operations compose in RKHS.

Perhaps the most practically significant result is our development of point-wise filter representations through the reproducing property (Theorems~\ref{thm_hom_box_prod} and~\ref{thm_alg_rkhs_vs_alg_classic}). We established that polynomial filters can be implemented via inner products $\langle q_v(K^{\Box}), f \rangle_{\mathcal{H}(K)}$, yielding computational advantages over classical implementations. Corollary~\ref{cor_diff_as_sum_Hkn} reveals that filtered signals decompose as sums of RKHS spaces induced by iterated box products, providing a bank filter interpretation where each term $h_r \boldsymbol{T}_K^r f$ lives in $\mathcal{H}(K^{\Box(r+1)})$. This enables efficient implementations when kernels $K^{\Box n}$ are precomputed.

The spatial-spectral localization results establish fundamental tradeoffs between frequency bandlimitation and RKHS-finite duration. Corollary~\ref{cor_spec_vs_ku} provides a direct relationship between filter coefficients and spectral response: $p(\sigma_i) = \frac{1}{\overline{\vartheta_i(v)}} \sum_{\ell \in \mathcal{L}} \alpha_\ell \overline{\vartheta_i(\ell)}$, revealing that the filter amplitude at frequency $\sigma_i$ is determined by a correlation between coefficient and eigenfunction sequences. Corollary~\ref{cor_rkhs_uncertainty} shows that RKHS-finite signals cannot be exactly bandlimited when $\sigma_i \neq 0$, but the eigenvalue decay enables approximate bandlimitation. The tradeoff is governed by $|\mathcal{T}| - B$: when $B < |\mathcal{T}|$, we allocate $B$ coefficients for low-pass behavior while using remaining coefficients to minimize residuals.

The representer theorem application in Corollary~\ref{cor_filt_representer} establishes rigorous foundations for learning convolutional filters in integral-operator-based neural architectures. When the spectrum $\{\sigma_i\}_{i=1}^\infty$ lies in $\mathcal{X}$, optimal filters admit finite-dimensional representations $p^*(u) = \sum_{i=1}^q a_i k_{\sigma_i}(u)$ as expansions with kernel functions centered at the eigenvalues. This enables a reformulation of filter learning in graphon neural networks as classical RKHS optimization problems, with Example~3 illustrating practical effectiveness for designing filters with prescribed spectral characteristics.

The digraphon extension in Theorem~\ref{thm_diff_digraphon} demonstrates framework robustness beyond symmetric operators. By showing $\boldsymbol{T}_K = \boldsymbol{T}_W \boldsymbol{T}_W^*$ for directed graphons, we established that $K = W \Box W$ provides well-defined spectral representation even without symmetry, enabling spectral analysis on directed networks through eigenvectors of the self-adjoint operator $\boldsymbol{T}_K$.

From a computational perspective, the point-wise representations in Theorems~\ref{thm_hom_box_prod} and~\ref{thm_alg_rkhs_vs_alg_classic} provide practical advantages: rather than iteratively applying integral operators $\boldsymbol{T}_K^r f$, one precomputes kernels $K^{\Box(r+1)}$ and evaluates filtered signals through RKHS inner products. This is particularly beneficial when multiple signals are filtered with the same polynomial, as computational cost is dominated by one-time kernel computation rather than repeated operator applications.

Several directions merit further investigation. The spatial-spectral localization results provide foundations for developing graphon uncertainty principles characterizing fundamental limits on joint localization. The connection between graphons and RKHS through Theorems~\ref{thm_K_from_W_gphonsp} and~\ref{thm_W_gphon_from_K} suggests graphon learning could be approached through kernel learning methodologies. The RKHS space decomposition in Corollary~\ref{cor_diff_as_sum_Hkn} opens possibilities for multi-resolution analysis in graphon signal processing, analogous to wavelet decompositions.

In conclusion, this paper establishes that integral-operator-based filtering naturally induces RKHS structures with well-defined algebraic properties. The box product algebra provides an intrinsic framework for polynomial filtering yielding point-wise filter representations through the reproducing property. Our results reveal deep connections between eigendecompositions and RKHS representations in graphon signal processing, extend to directed graphons, and enable spatial-spectral localization analyses. The representer theorem formulation provides principled foundations for learnable convolutional architectures based on integral operators. By demonstrating equivalence between classical operator-based filtering and RKHS-based point-wise implementations, we have provided both theoretical insights and computational tools for analyzing signals on continuous network models, complementing and extending algebraic signal processing theory.




\bibliography{bibliography}

@misc{algSP0,
    title={Algebraic Signal Processing Theory},
    author={Markus P{\"u}schel and Jos{\'e} M. F. Moura},
    year={2006},
    eprint={cs/0612077},
    archivePrefix={arXiv},
    primaryClass={cs.IT}
}

@ARTICLE{algSP1,
AUTHOR = {Markus P{\"u}schel and Jos{\'e} M. F. Moura},
TITLE = {Algebraic Signal Processing Theory: Foundation and {1-D} Time},
JOURNAL = {IEEE Transactions on Signal Processing},
VOLUME = {56},
NUMBER = {8},
PAGES = {3572--3585},
YEAR = {2008}
}

@ARTICLE{algSP2,
AUTHOR = {Markus P{\"u}schel and Jos{\'e} M. F. Moura},
TITLE = {Algebraic Signal Processing Theory: {1-D} Space},
JOURNAL = {IEEE Transactions on Signal Processing},
VOLUME = {56},
NUMBER = {8},
PAGES = {3586--3599},
YEAR = {2008}
}

@ARTICLE{algSP6,
AUTHOR = {Markus P{\"u}schel and Martin R\"{o}tteler},
TITLE = {Algebraic Signal Processing Theory: {2-D} Spatial Hexagonal Lattice},
JOURNAL = {IEEE Transactions on Image Processing},
VOLUME = {16},
NUMBER = {6},
PAGES = {1506--1521},
YEAR = {2007}
}

@ARTICLE{algSP7,
AUTHOR = {Aliaksei Sandryhaila and Jelena Kovacevic and Markus P{\"u}schel},
TITLE = {Algebraic Signal Processing Theory: {1-D} Nearest-Neighbor Models},
JOURNAL = {IEEE Transactions on Signal Processing},
VOLUME = {60},
NUMBER = {5},
PAGES = {2247--2259},
YEAR = {2012}
}

@ARTICLE{algSP8,
AUTHOR = {Aliaksei Sandryhaila and Jelena Kovacevic and Markus P{\"u}schel},
TITLE = {Algebraic Signal Processing Theory: {Cooley}-{Tukey} Type Algorithms for Polynomial Transforms Based on Induction},
JOURNAL = {SIAM Journal on Matrix Analysis and Applications},
VOLUME = {32},
NUMBER = {2},
PAGES = {364--384},
YEAR = {2011}
}

@ARTICLE{puschel_asplattice,
	author={P{\"u}schel, Markus and Seifert, Bastian and Wendler, Chris},
	journal={IEEE Transactions on Signal Processing}, 
	title={Discrete Signal Processing on Meet/Join Lattices}, 
	year={2021},
	volume={69},
	number={},
	pages={3571-3584},
	doi={10.1109/TSP.2021.3081036}}

@ARTICLE{puschel_aspsets,
	author={P{\"u}schel, Markus and Wendler, Chris},
	journal={IEEE Transactions on Signal Processing}, 
	title={Discrete Signal Processing with Set Functions}, 
	year={2021},
	volume={69},
	number={},
	pages={1039-1053},
	doi={10.1109/TSP.2020.3046972}}

@ARTICLE{puschel_digraphs1,
	author={Seifert, Bastian and Wendler, Chris and Püschel, Markus},
	journal={IEEE Transactions on Signal Processing}, 
	title={Causal Fourier Analysis on Directed Acyclic Graphs and Posets}, 
	year={2023},
	volume={71},
	number={},
	pages={3805-3820},
	doi={10.1109/TSP.2023.3324988}}

@INPROCEEDINGS{puschel_digraphs2,
	author={Mihal, Vedran and Püschel, Markus},
	booktitle={ICASSP 2023 - 2023 IEEE International Conference on Acoustics, Speech and Signal Processing (ICASSP)}, 
	title={Möbius Total Variation for Directed Acyclic Graphs}, 
	year={2023},
	volume={},
	number={},
	pages={1-5},
	doi={10.1109/ICASSP49357.2023.10095435}}

@book{paulsen2016introduction,
  title={An Introduction to the Theory of Reproducing Kernel Hilbert Spaces},
  author={Paulsen, V.I. and Raghupathi, M.},
  isbn={9781316558737},
  series={Cambridge Studies in Advanced Mathematics},
  url={https://books.google.com/books?id=QDvzCwAAQBAJ},
  year={2016},
  publisher={Cambridge University Press}
}

@book{wainwright2019high,
  title={High-Dimensional Statistics: A Non-Asymptotic Viewpoint},
  author={Wainwright, M.J.},
  isbn={9781108498029},
  lccn={2018043475},
  series={Cambridge Series in Statistical and Probabilistic Mathematics},
  url={https://books.google.com/books?id=IluHDwAAQBAJ},
  year={2019},
  publisher={Cambridge University Press}
}

@article{ghojogh2021reproducing,
	title={Reproducing Kernel Hilbert Space, Mercer's Theorem, Eigenfunctions, Nystr$\backslash$" om Method, and Use of Kernels in Machine Learning: Tutorial and Survey},
	author={Ghojogh, Benyamin and Ghodsi, Ali and Karray, Fakhri and Crowley, Mark},
	journal={arXiv preprint arXiv:2106.08443},
	year={2021}
}

@book{halmos2012bounded,
	title={Bounded Integral Operators on L 2 Spaces},
	author={Halmos, P.R. and Sunder, V.S.},
	isbn={9783642670169},
	series={Ergebnisse der Mathematik und ihrer Grenzgebiete. 2. Folge},
	url={https://books.google.com/books?id=XfQqBAAAQBAJ},
	year={2012},
	publisher={Springer Berlin Heidelberg}
}

@INPROCEEDINGS{lga_icassp,
author={Kumar, Harshat and Parada-Mayorga, Alejandro and Ribeiro, Alejandro},
booktitle={ICASSP 2023 - 2023 IEEE International Conference on Acoustics, Speech and Signal Processing (ICASSP)}, 
title={Algebraic Convolutional Filters on Lie Group Algebras}, 
year={2023},
volume={},
number={},
pages={1-5},
doi={10.1109/ICASSP49357.2023.10095164}}

@article{lga_j,
	title={Lie Group Algebra Convolutional Filters},
	author={Kumar, Harshat and Parada-Mayorga, Alejandro and Ribeiro, Alejandro},
	journal={arXiv preprint arXiv:2305.04431},
	year={2023}
}

@INPROCEEDINGS{gphon_pooling_c,
  author={Parada-Mayorga, Alejandro and Ruiz, Luana and Ribeiro, Alejandro},
  booktitle={2020 28th European Signal Processing Conference (EUSIPCO)}, 
  title={Graphon Pooling in Graph Neural Networks}, 
  year={2021},
  volume={},
  number={},
  pages={860-864},
  doi={10.23919/Eusipco47968.2020.9287735}}

@ARTICLE{gphon_pooling_j,
  author={Parada-Mayorga, Alejandro and Wang, Zhiyang and Ribeiro, Alejandro},
  journal={IEEE Transactions on Signal Processing}, 
  title={Graphon Pooling for Reducing Dimensionality of Signals and Convolutional Operators on Graphs}, 
  year={2023},
  volume={71},
  number={},
  pages={3577-3591},
  doi={10.1109/TSP.2023.3318471}}

@ARTICLE{gphon_samp,
	author={Parada-Mayorga, Alejandro and Ribeiro, Alejandro},
	journal={IEEE Transactions on Signal Processing}, 
	title={Sampling and Uniqueness Sets in Graphon Signal Processing}, 
	year={2025},
	volume={73},
	number={},
	pages={2480-2495},
	doi={10.1109/TSP.2025.3577112}}

@ARTICLE{rkhs_conv,
	author={Parada-Mayorga, Alejandro and Agorio, Leopoldo and Ribeiro, Alejandro and Bazerque, Juan},
	journal={IEEE Transactions on Signal Processing}, 
	title={Convolutional Filtering With RKHS Algebras}, 
	year={2025},
	volume={73},
	number={},
	pages={2353-2367},
	doi={10.1109/TSP.2025.3572861}}

@ARTICLE{algnn_nc_j,
author={Parada-Mayorga, Alejandro and Butler, Landon and Ribeiro, Alejandro},
journal={IEEE Transactions on Signal Processing}, 
title={Convolutional Filters and Neural Networks With Noncommutative Algebras}, 
year={2023},
volume={71},
number={},
pages={2683-2698},
doi={10.1109/TSP.2023.3293716}}

@ARTICLE{parada_algnn,
author={Parada-Mayorga, A. and Ribeiro, A.},
journal={IEEE Transactions on Signal Processing}, 
title={Algebraic Neural Networks: Stability to Deformations}, 
year={2021},
volume={69},
number={},
pages={3351-3366},
doi={10.1109/TSP.2021.3084537}}

@INPROCEEDINGS{parada_algnnconf,
author={Parada-Mayorga, A. and Ribeiro, A.},
booktitle={ICASSP 2021 - 2021 IEEE International Conference on Acoustics, Speech and Signal Processing (ICASSP)}, 
title={Stability of Algebraic Neural Networks to Small Perturbations}, 
year={2021},
volume={},
number={},
pages={5205-5209},
doi={10.1109/ICASSP39728.2021.9414604}}

@article{parada_quiversp,
	title={Quiver Signal Processing (QSP)},
	author={A. {Parada-Mayorga} and H. {Riess} and A. {Ribeiro} and R. {Ghrist}},
	journal={ArXiv},
	year={2020},
	volume={abs/2010.11525}
}

@article{msp_j,
	title={Convolutional learning on multigraphs},
	author={Butler, Landon and Parada-Mayorga, Alejandro and Ribeiro, Alejandro},
	journal={IEEE Transactions on Signal Processing},
	volume={71},
	pages={933--946},
	year={2023},
	publisher={IEEE}
}

@INPROCEEDINGS{msp_icassp2023,
	author={Butler, Landon and Parada-Mayorga, Alejandro and Ribeiro, Alejandro},
	booktitle={ICASSP 2023 - 2023 IEEE International Conference on Acoustics, Speech and Signal Processing (ICASSP)}, 
	title={Learning with Multigraph Convolutional Filters}, 
	year={2023},
	volume={},
	number={},
	pages={1-5},
	doi={10.1109/ICASSP49357.2023.10095028}}

@ARTICLE{gphon_leus,
	author={Morency, Matthew W. and Leus, Geert},
	journal={IEEE Transactions on Signal Processing}, 
	title={Graphon Filters: Graph Signal Processing in the Limit}, 
	year={2021},
	volume={69},
	number={},
	pages={1740-1754},
	doi={10.1109/TSP.2021.3061575}}

@article{diao2016model,
	title={Model-free consistency of graph partitioning},
	author={Diao, Peter and Guillot, Dominique and Khare, Apoorva and Rajaratnam, Bala},
	journal={arXiv preprint arXiv:1608.03860},
	year={2016}
}

@book{lovaz2012large,
  title={Large Networks and Graph Limits},
  author={Lov{\'a}sz, L.},
  isbn={9780821890851},
  lccn={2012034211},
  series={American Mathematical Society colloquium publications},
  url={https://books.google.com/books?id=FsFqHLid8sAC},
  year={2012},
  publisher={American Mathematical Society}
}

@article{digraphon1,
	title={Convergence of spectra of digraph limits},
	author={Greb{\'\i}k, Jan and Kr{\'a}l, Daniel and Liu, Xizhi and Pikhurko, Oleg and Slipantschuk, Julia},
	journal={arXiv preprint arXiv:2506.04426},
	year={2025}
}

@ARTICLE{digraphon2,
	author={Fabian, Christian and Cui, Kai and Koeppl, Heinz},
	journal={IEEE Control Systems Letters}, 
	title={Mean Field Games on Weighted and Directed Graphs via Colored Digraphons}, 
	year={2023},
	volume={7},
	number={},
	pages={877-882},
	doi={10.1109/LCSYS.2022.3227453}}

@article{digraphon3,
	title={Efficient Evolutionary Models with Digraphons},
	author={Tamaskar, Abhinav and Mishra, Bud},
	journal={arXiv preprint arXiv:2104.12748},
	year={2021}
}

@book{conway1994course,
  title={A Course in Functional Analysis},
  author={Conway, J.B.},
  isbn={9780387972459},
  lccn={97122669},
  series={Graduate Texts in Mathematics},
  url={https://books.google.com/books?id=ix4P1e6AkeIC},
  year={1994},
  publisher={Springer New York}
}

@book{aliprantis2002invitation,
  title={An Invitation to Operator Theory},
  author={Aliprantis, C.D.},
  isbn={9780821872291},
  series={Graduate studies in mathematics},
  url={https://books.google.com/books?id=5h8bEWNalwMC},
  year={2002},
  publisher={American Mathematical Society}
}
\bibliographystyle{unsrt}

\ifCLASSOPTIONcaptionsoff
  \newpage
\fi


\clearpage
\newpage
\begin{center}
{\LARGE \textbf{Supplementary Material}}\\[1em]
{\large Proofs and Derivations}
\end{center}
\vspace{1em}


\appendices

\section{Proofs and Extra Theorems}


\subsection{Proof of Theorem~\ref{thm_K_from_W_gphonsp}}
\label{proof_thm_K_from_W_gphonsp}

\begin{proof}
From Theorem~11.3 in~\cite{paulsen2016introduction} it follows that $\ccalH$ is an RKHS with kernel given by $K=W\square W$. Additionally, it follows that
\begin{multline}
\boldsymbol{T}_{K}
         f 
            =
                \int_{0}^{1}K(u,v)f(v)dv
                    =
                    \\
                       \int_{0}^{1}
                            \left(
                                   \int_{0}^{1}
                                        W(u,z)
                                        W(z,v)
                                        dz
                            \right)
                                    f(v)dv
                                    .
\end{multline}
Using Fubini's theorem for Lebesgue integrable functions on $[0,1]^2$ and rearranging the terms, we have
\begin{equation}
\boldsymbol{T}_{K}
         f 
           =
               \int_{0}^{1}W(u,z)
                      \left(
                              \int_{0}^{1}W(z,v)f(v)dv
                      \right) 
                      dz
                      .      
\end{equation}
Then, from the definition of the $\boldsymbol{T}_{W}$ it follows that
\begin{equation}
\boldsymbol{T}_{K}
         f
          =
            \boldsymbol{T}_{W}\left(
                                                  \boldsymbol{T}_{W}f
                                            \right)
          =
            \boldsymbol{T}_{W}^{2}f    
            .                             
\end{equation}

If $\varphi_{i}$ is the $i$-th eigenvector of $\boldsymbol{T}_{W}$ with eigenvalue $\lambda_{i}$, then it follows that
\begin{equation}
\boldsymbol{T}_{K}
         \varphi_{i}
          =
            \boldsymbol{T}_{W}^{2}\varphi_{i}   
          =  
            \lambda_{i}^{2}\varphi_{i}
            .     
\end{equation}
This means that $\varphi_{i}$ is an eigenvector of $\boldsymbol{T}_{K}$ with eigenvalue $\lambda_{i}^{2}$.

\end{proof}


\subsection{Proof of Theorem~\ref{thm_W_gphon_from_K}}
\label{proof_thm_W_gphon_from_K}

\begin{proof}
Since $K(u,v)\geq 0$ is continuous on the compact set $[0,1]^{2}$, $W(u,v)=K(u,v)/C$ is bounded an measurable. Additionally, the constant $C$ guarantees that $0\leq W(u,v)\leq 1$. Additionally, as per the properties of $K$ as a reproducing kernel and the fact that $K(u,v)\geq 0$ we have $K(u,v)=K(v,u)$ which ensures $W(u,v)$ is symmetric.
\end{proof}


\subsection{Proof of Theorem~\ref{thm_gphon_fourier_kv}}
\label{proof_thm_gphon_fourier_kv}

\begin{proof}
From Theorem~\ref{thm_K_from_W_gphonsp} we ensure that the eigenvalues and eigenvectors of $\boldsymbol{T}_{K}$ are $\{ \varphi_{i}(u) \}_{i=1}^{\infty}$ and $\{ \lambda_{i} \}_{i=1}^{\infty}$, respectively. Then, using~\eqref{eq_mercer_K_decomp} it follows that
$
K(u,v)
     =
      \sum_{i=1}^{\infty}
            \lambda_{i}^{2}
                    \varphi_{i}(u)
                    \varphi_{i}(v)
                    .
$
Then, expressing $k_{v}(u)$ as an expansion in terms of the $\{ \varphi_{i}(u)\}_{i=1}^{\infty}$ we have
\begin{equation}\label{eq_proof_thm_gphon_fourier_kv_1}
k_{v}(u) = 
           \sum_{i=1}^{\infty}
                    \left(
                         \lambda_{i}^{2}
                         \varphi_{i}(v)
                    \right)
                    \varphi_{i}(u)
                    ,
\end{equation}
which shows that the Fourier coefficients of $k_{v}(u)$ are given by~\eqref{eq_thm_gphon_fourier_kv_1}. Additionally, since $f=\sum_{v\in [0,1]\alpha_{v}k_{v}(u)}$, \eqref{eq_thm_gphon_fourier_kv_2} follows from combining the expansion of $f$ in terms of $k_{v}(u)$ with~\eqref{eq_proof_thm_gphon_fourier_kv_1}.
\end{proof}


\subsection{Proof of Theorem~\ref{thm_diff_digraphon}}
\label{proof_thm_diff_digraphon}

\begin{proof}
Since any digraphon $W:[0,1]^{2}\to [0,1]$ is by definition bounded and measurable, then by Theorem~11.3 in~\cite{paulsen2016introduction} we can ensure that
$
\ccalH 
        =
          \left\lbrace
                \left. 
                      \boldsymbol{T}_{W}f
                \right\vert
                             f\in L_{2}[0,1]      
          \right\rbrace
$
is an RKHS with a reproducing kernel given by $K=W\square W$. Additionally, since $[0,1]$ is compact and bounded, $\boldsymbol{T}_{K}$ has an spectral decomposition given according to~\eqref{eq_TK_operator} and~\eqref{eq_mercer_K_decomp}. Now, we recall that
\begin{multline}
\boldsymbol{T}_{K}
         f 
            =
                \int_{0}^{1}K(u,v)f(v)dv
                    =
                    \\
                       \int_{0}^{1}
                            \left(
                                   \int_{0}^{1}
                                        W(u,z)
                                        W(v,z)
                                        dz
                            \right)
                                    f(v)dv
                                    .
\end{multline}
Then, by Fubini's theorem and regrouping terms we obtain
\begin{multline}
\int_{0}^{1}
     \left(
         \int_{0}^{1}
                W(u,z)
                W(v,z)
                dz
        \right)
              f(v)dv  
              =
              \\
               \int_{0}^{1}
                    W(u,z)
                    \left( 
                        \int_{0}^{1}
                            W(v,z)
                            f(v)
                            dv
                    \right)dz
                    .
\end{multline}
Then, taking into account the definition of $\boldsymbol{T}_{W^{\ast}}$ and $\boldsymbol{T}_{W}$ we conclude that 
\begin{equation}
\boldsymbol{T}_{K}
      =
       \boldsymbol{T}_{W}\boldsymbol{T}_{W^{*}}
       .
\end{equation}

\end{proof}


\subsection{Proof of Theorem~\ref{thm_alg_RKHS_box_prod_1}}
\label{proof_thm_alg_RKHS_box_prod_1}

\begin{proof}
We verify that $\ccalA_{K}$ is an algebra. We start verifying that $\ccalA_{K}$ is indeed a vector space. First, we verify that the sum  in $\ccalA_{K}$ is closed. This follows trivially from
\begin{equation}
    \sum_{r=0}^{\infty}a_{r}K^{\square r}
+
    \sum_{r=0}^{\infty}b_{r}K^{\square r}
     =
     \sum_{r=0}^{\infty}
              \left( 
                     a_{r}+b_{r}
             \right)        
           K^{\square r}
           .
\end{equation}

Now, we verify that the scalar multiplication is closed in $\ccalA_{K}$, which follows from
\begin{equation}
\alpha
       \sum_{r=0}^{\infty}a_{r}K^{\square r}
               =
                 \sum_{r=0}^{\infty}
                                 \left( 
                                        \alpha a_{r}
                                 \right)
                                          K^{\square r}   
                                          .   
\end{equation}

Now, we verify that ``$\bullet$" is an algebra product. First, we note that the closed-ness of $\bullet$ follows trivially from~\eqref{eq_A_rkhs_box_2}. Regarding the associativity of $\bullet$, we must verify that 
$
\left(
       p_{1}\left( K^{\square} \right)
               \bullet
        p_{2}\left( K^{\square} \right)
\right)
      \bullet 
      p_{3}\left( K^{\square} \right)
=
p_{1}\left( K^{\square} \right)\bullet
\left(
        p_{2}\left( K^{\square} \right)
                \bullet
        p_{3}\left( K^{\square} \right)
\right)
,
$
for all $ p_{1}\left( K^{\square} \right), p_{2}\left( K^{\square} \right), p_{3}\left( K^{\square} \right)\in\ccalA_{K}$. Now, let us consider
\begin{multline}
p_{1}\left( K^{\square} \right)
=
\left( 
    \sum_{r=0}^{\infty}x_{r}K^{\square r}
\right)
,
~
p_{2}\left( K^{\square} \right)
=
\left( 
    \sum_{r=0}^{\infty}y_{r}K^{\square r}
\right)
,
\\
p_{3}\left( K^{\square} \right)
=
\left( 
     \sum_{r=0}^{\infty}z_{r}K^{\square r}
\right)
.
\end{multline}

Then, we start computing
\begin{equation}
\left(
p_{1}\bullet p_{2}
\right)
     \bullet p_{3}
     =
\sum_{r,\ell,m=0}^{\infty}
         \left(
             x_{r}y_{\ell}
         \right)z_{m}
         \left( 
               K^{\square (r+\ell)}
         \right)
                 K^{\square m}
 .
\end{equation}
Since the scalar product and the sum are associative it then follows that
\begin{multline}
\left(
p_{1}\bullet p_{2}
\right)
     \bullet p_{3}
     =
\sum_{r,\ell,m=0}^{\infty}
             x_{r}
         \left(    
             y_{\ell}
             z_{m}
         \right)
               K^{\square r}
         \left( 
               K^{\square (\ell+m)}
         \right)
=  
\\
p_{1}\bullet
\left(
      p_{2}
       \bullet
      p_{3}
\right)           
 .
\end{multline}
With the associativity property at hand we proceed to show the distributive property of $\bullet$ with respect to the sum in $\ccalA_{K}$. To this end we start with
\begin{equation}
(p_{1}+p_{2})\bullet p_{3}
     =
      \left( 
           \sum_{r=0}^{\infty}
                (x_{r}+y_{r})K^{\square r}
      \right)
      \bullet 
      \sum_{\ell=0}^{\infty}
      z_{\ell}K^{\square \ell}
      .
\end{equation}
Then, by~\eqref{eq_A_rkhs_box_2} this leads to
\begin{equation}
      \left( 
           \sum_{r=0}^{\infty}
                (x_{r}+y_{r})K^{\square r}
      \right)
      \bullet 
      \sum_{\ell=0}^{\infty}
      z_{\ell}K^{\square \ell}
      =
      \sum_{r,\ell=0}^{\infty}
          (x_{r}+y_{r})z_{\ell}
          K^{\square (r+\ell)}
          .
\end{equation}
Since the ordinary scalar products and and sum satisfy the distributive property, it follows that
\begin{equation}
\sum_{r,\ell=0}^{\infty}
          (x_{r}+y_{r})z_{\ell}
          K^{\square (r+\ell)}
      =
      \sum_{r,\ell=0}^{\infty}
          x_{r}z_{\ell}K^{\square (r+\ell)}
          +
      \sum_{r,\ell=0}^{\infty}    
          y_{r}z_{\ell}K^{\square (r+\ell)}
          .
\end{equation}
Finally, by definition of $\bullet$ in~\eqref{eq_A_rkhs_box_2} we have
\begin{equation}
\sum_{r,\ell=0}^{\infty}
          x_{r}z_{\ell}K^{\square (r+\ell)}
          +
\sum_{r,\ell=0}^{\infty}    
          y_{r}z_{\ell}K^{\square (r+\ell)}   
          =
           p_{1}\bullet p_{3}
           +
           p_{2}\bullet p_{3}
           .
\end{equation}
\end{proof}


\subsection{Proof of Theorem~\ref{thm_box_prod_spect}}
\label{proof_thm_box_prod_spect}

\begin{proof}
The box product between $S_{1}$ and $S_{2}$ is given by
\begin{multline}
\left(
    S_{1}\square S_{2}
\right)
       =
        \int_{\ccalX}
                  S_{1}(u,z)
                  S_{2}(z,v)d\mu(z)
       \\           
       =
        \int_{\ccalX}
                 \left( 
                         \sum_{i=1}^{\infty}
                                   a_{i}
                                   \vartheta_{i}(u)
                                   \overline{\vartheta_{i}(z)}
                 \right)    
                 \left( 
                         \sum_{j=1}^{\infty}
                                   b_{j}
                                   \vartheta_{j}(z)
                                   \overline{\vartheta_{j}(v)}
                 \right)   
                 d\mu (z)     
                 .   
\end{multline}
Then, distributing the sum and rearranging terms we have
\begin{multline}
\int_{\ccalX}
                  S_{1}(u,z)
                  S_{2}(z,v)d\mu(z)
                  \\
        =          
\sum_{i,j=1}^{\infty}
         a_{i}b_{j}
             \vartheta_{i}(u)
             \overline{\vartheta_{j}(v)}
                  \int_{\ccalX}
                      \overline{\vartheta_{i}(z)}
                      \vartheta_{j}(z) 
                      d\mu (z)     
                      .   
\end{multline}
Since the $\vartheta_{i}$ constitute an orthonormal basis in $L_{2}(\ccalX, \mu)$ we have 
\begin{equation}
\int_{\ccalT}
                  S_{1}(u,z)
                  S_{2}(z,v)d\mu(z)
                  =
                  \sum_{i=1}^{\infty}
                           a_{i}
                           b_{i}
                              \vartheta_{i}(u)
                              \overline{\vartheta_{i}(v)}
                           .
\end{equation}
From this, it follows trivially that
\begin{equation}
S^{n\square}
        =
        \sum_{i=1}^{\infty}
                \lambda_{i}^{n}
                          \vartheta_{i}(u)
                          \overline{\vartheta_{i}(v)}
                          ,
\end{equation}
for any $n\in\mbN$. Then, for any polynomial $p(t)=\sum_{r=0}^{R}h_{r}t^{r}$ we obtain
\begin{equation}
 p\left( 
      S^{\square}
  \right)
        =\sum_{r=0}^{R}
              h_{r}
                 \left( 
                       \sum_{i=1}^{\infty}
                               \lambda_{i}^{r}
                               \vartheta_{i}(u)
                               \overline{\vartheta_{i}(v)}
                 \right)
                 .
\end{equation}
Rearranging the sums we obtain
\begin{multline}
p\left( 
      S^{\square}
  \right)
  =
\sum_{i=1}^{\infty}
          \left( 
                \sum_{r=0}^{R}
                     h_{r}
                     \lambda_{i}^{r}
          \right)
           \vartheta_{i}(u)
           \overline{\vartheta_{i}(v)}
           \\
           =
       \sum_{i=1}^{\infty}
             p\left( 
                 \lambda_{i}
              \right)
              \vartheta_{i}(u)
              \overline{\vartheta_{i}(v)}           
           .
\end{multline}
\end{proof}


\subsection{Proof of Theorem~\ref{thm_hom_box_prod}}
\label{sub_sec_proof_thm_Ak_subsets_in_HK}

\begin{proof}

First, we proceed to prove that $q_{v}\left( K^{\square}\right)\in\ccalH(K) $. To this end we must show that
\begin{equation}\label{eq_sub_sec_proof_thm_Ak_subsets_in_HK_1}
\sum_{j=1}^{\infty}
        \frac{1}{\sigma_{j}}
             \left\vert
                  \left\langle 
                       q_{v}\left(K^{\square}\right)
                       ,
                        \vartheta_{j}
                   \right\rangle_{L_{2}}
            \right\vert^{2}
             <
             \infty
             .
\end{equation}

We start taking into account that in the light of Theorem~\ref{thm_box_prod_spect} we have
\begin{equation}\label{eq_sub_sec_proof_thm_Ak_subsets_in_HK_2}
q_{v}\left(K^{\square}\right)
=
\left(
p 
    \left(
             K^{\square}
    \right) 
    \square
    K
\right)(u,v)  
    =    
       \sum_{j=1}^{\infty}     
            p\left( 
                    \sigma_{j}
              \right)    
              \sigma_{j}      
              \vartheta_{j}(u)
              \overline{\vartheta_{j}(v)}
              .
\end{equation}
Then, it follows that
\begin{equation}
\sum_{i=1}^{\infty}
        \frac{1}{\sigma_{i}}
             \left\vert
             \left\langle 
                   q_{v}\left(K^{\square}\right)
                   ,
                   \vartheta_{i}
             \right\rangle_{L_{2}}
             \right\vert^{2}
    =
    \sum_{i=1}^{\infty}  
                           p\left( 
                                    \sigma_{i}
                              \right)^{2} 
                              \sigma_{i}
                              \left\vert 
                                    \vartheta_{i}(v)
                              \right\vert^{2}  
             .
\end{equation}
Then, using the H\"older inequality we obtain 
\begin{multline}
\sum_{i=1}^{\infty}  
                           p\left( 
                                    \sigma_{i}
                              \right)^{2} 
                           \sigma_{i}   
                           \left\vert
                                \vartheta_{i}(v)
                           \right\vert^{2}                            
\leq 
                 \left( 
                        \max_{i}
                               p(\sigma_i)^{2}
                 \right)     
                 \left( 
                        \sum_{i=1}^{\infty}
                                  \sigma_{i}       
                                  \left\vert 
                                       \vartheta_{i}(v)
                                  \right\vert^{2}  
                 \right)     
                 .         
\end{multline}
With this result at hand, we emphasize some observations. First, we notice that 
$
 \sum_{i=1}^{\infty}
                      \sigma_{i}       
                      \left\vert 
                           \vartheta_{i}(v)
                      \right\vert^{2}  
=
K(v,v)  
,                    
$
and since $K$ is continuous, there exists $0<C<\infty$ such that $\vert K(u,v)\vert\leq C$ for all $u,v\in\ccalX$~\cite{paulsen2016introduction}. Second, notice that $T_{K}$ is a bounded operator, therefore the spectrum of $T_{K}$ is always bounded~\cite{conway1994course,aliprantis2002invitation}. This ensures that there exists $c>0$ such that $0\leq\sigma_{j}<c$. Then, if $p(t)$ is a polynomial we have that for any $t\in [0,c]$ there exists $D>0$ such that $p(t)^{2}<D$. Putting this together we have that
$
\sum_{i=1}^{\infty}  
                           p\left( 
                                    \sigma_{i}
                              \right)^{2} 
                           \sigma_{i}   
                           \left\vert 
                                \vartheta_{i}(v)
                           \right\vert^{2}   
<
   \infty
   ,
$
which implies that $q_{v}\left(K^{\square}\right)\in\ccalH(K)$.

Now, we show that $g(v)=\overline{\left\langle q_{v}\left(  K^{\square} \right),f\right\rangle}_{\ccalH(K)}\in\ccalH(K)$. To this end, we have to show that 
\begin{equation}\label{eq_sub_sec_proof_thm_Ak_subsets_in_HK_1}
\sum_{j=1}^{\infty}
        \frac{1}{\sigma_{j}}
             \left\vert 
             \left\langle
                   \overline{\left\langle
                       q_{v}\left(K^{\square}\right)
                       ,
                       f
                   \right\rangle}_{\ccalH(K)}    
                   ,
                   \vartheta_{j}(v)
             \right\rangle_{L_{2}}
             \right\vert^{2}
             <
             \infty
             .
\end{equation}
We start leveraging~\eqref{eq_sub_sec_proof_thm_Ak_subsets_in_HK_2} to obtain  
\begin{multline}
\sum_{j=1}^{\infty}
            \frac{1}{\sigma_j}
             \left\langle
                             \sum_{i=1}^{\infty}     
                                           p\left( 
                                                   \sigma_{i}
                                             \right) 
                                             \sigma_{i}         
                                             \vartheta_{i}(v)
                                             \langle
                                                       f
                                                       ,
                                                       \vartheta_{i}
                                             \rangle_{\ccalH(K)}
                     ,
                     \vartheta_{j}(v)
             \right\rangle_{L_{2}}^{2}
             \\
             =
\sum_{j=1}^{\infty}  
                           p\left( 
                                    \sigma_{j}
                              \right)^{2} 
                              \sigma_{j}
                              \langle
                                        f
                                        ,
                                        \vartheta_{j}
                             \rangle_{\ccalH(K)}^{2}
                           . 
\end{multline}  
By using the H\"older inequality it follows that
\begin{multline}
\sum_{j=1}^{\infty}  
                           p\left( 
                                    \sigma_{j}
                              \right)^{2} 
                              \sigma_{j}
                              \langle
                                        f
                                        ,
                                        \vartheta_{j}
                             \rangle_{\ccalH(K)}^{2}
                     \leq 
                     \\
                           \left( 
                                  \sum_{j=1}^{\infty}  
                                           \sigma_{j}
                                           \langle
                                                  f
                                                  ,
                                                  \vartheta_{i}
                                           \rangle_{\ccalH(K)}^{2}
                           \right)    
                           \max_{j}
                           p(\sigma_{j})^{2}  
                           .
\end{multline}
Since $\boldsymbol{T}_{K}$ is a bounded operator, there exists $c>0$ such that $0\leq\sigma_{j}<c$~\cite{conway1994course,aliprantis2002invitation}. Then, if $p(t)$ is a polynomial we have that for any $t\in [0,c]$ there exists $D>0$ such that $p(t)^{2}<D$. 

If we take into account that
\begin{equation}
\left\langle
            f,\vartheta_{i}
\right\rangle_{\ccalH(K)}
          =
            \left\vert 
                  \left\langle f,\vartheta_{i}\right\rangle_{\ccalH(K)}
            \right\vert e^{j\theta}   
            ,
\end{equation}
for some $\theta$, by the H\"older inequality we obtain       
\begin{equation}\label{eq_sub_sec_proof_thm_Ak_subsets_in_HK_aux_1}
\sum_{i=1}^{\infty}
        \sigma_{i}        
        \left\langle 
                f
                ,
                \vartheta_{i}
        \right\rangle_{\ccalH(K)}^{2}
        =
          \sum_{i=1}^{\infty}
                 \sigma_{i}        
                 \left\vert \left\langle 
                f
                ,
                \vartheta_{i}
        \right\rangle_{\ccalH(K)}\right\vert^{2}
        e^{j2\theta}
.   
\end{equation}
Now, taking into account that
\begin{equation}
\sum_{i=1}^{\infty}
        \sigma_{i} 
        \left\vert
        \left\langle 
                f
                ,
                \vartheta_{i}
        \right\rangle_{\ccalH(K)}
        \right\vert^{2}
         =
        \left\langle 
                \boldsymbol{T}_{K}f
                ,
                \boldsymbol{T}_{K}f
        \right\rangle_{\ccalH(K)}   
        =
        \left\Vert 
               \boldsymbol{T}_{K}f
        \right\Vert_{\ccalH(K)}^{2} 
        <
        \infty
,   
\end{equation}
and the Holder inequality in~\eqref{eq_sub_sec_proof_thm_Ak_subsets_in_HK_aux_1} we have that
$
\sum_{j=1}^{\infty}  
                           p\left( 
                                    \sigma_{j}
                              \right)^{2} 
                              \sigma_{j}
                              \langle
                                      f
                                      ,
                                      \vartheta_{j}
                              \rangle_{\ccalH(K)}^{2}
<
\infty
.
$

Now, we proceed to show that the map $\rho_{K}$ is a linear map that preserves the products in the algebra. The linearity of $\rho_{K}$ follows trivially from the linearity of the inner product $\langle\cdot,\cdot\rangle_{\ccalH(K)}$. Then we focus our attention in showing that $\rho_{K}$ preserves the product. We start taking into account that
\begin{multline}
\rho_{K}
     \left( 
            p(t)
     \right)
             \left(
                      \rho_{K}
                             \left( 
                                   r(t)
                              \right)
                                    f
             \right)
=
\\
\overline{
\left\langle
       \sum_{j=1}^{\infty}
                p(\sigma_{j})
                       \sigma_{j}
                       \vartheta_{j}(u)
                       \overline{\vartheta_{j}(v)}
        ,
        \sum_{i=1}^{\infty}
                r(\sigma_{i})
                       \sigma_{i}
                        \vartheta_{i}(u)
                        \overline{
                        \left\langle
                               \vartheta_{i}
                               ,
                               f
                        \right\rangle}_{\ccalH(K)}
\right\rangle}_{\ccalH(K)}    
.        
\end{multline}
Developing the inner product we obtain
\begin{multline}
\rho_{K}
     \left( 
            p(t)
     \right)
             \left(
                      \rho_{K}
                             \left( 
                                   r(t)
                              \right)
                                    f
             \right)
=
\\
\overline{
\sum_{i=1}^{\infty}
         p(\sigma_{i})
         r(\sigma_{i})
         \sigma_{i}^{2}
         \overline{\vartheta_{j}(v)}
         \langle
               \vartheta_{i}
               ,
               f
         \rangle_{\ccalH(K)}
         \langle
                \vartheta_{i}
                ,
                \vartheta_{i}
         \rangle_{\ccalH(K)}
}  
.        
\end{multline}
Then, taking into account that $\langle \vartheta_i, \vartheta_i\rangle_{\ccalH(K)}=1/\sigma_i$ and developing the conjugate operation it follows that
\begin{multline}
\rho_{K}
     \left( 
            p(t)
     \right)
             \left(
                      \rho_{K}
                             \left( 
                                   r(t)
                              \right)
                                    f
             \right)
=
\\
\sum_{i=1}^{\infty}
         p(\sigma_{i})
         r(\sigma_{i})
         \sigma_{i}
         \vartheta_{j}(v)
         \overline{
         \langle
               \vartheta_{i}
               ,
               f
         \rangle}_{\ccalH(K)}
         \\
=       
\sum_{i=1}^{\infty}
         p(\sigma_{i})
         r(\sigma_{i})
         \sigma_{i}
         \vartheta_{j}(v)
         \langle
               f
               ,
               \vartheta_{i}
         \rangle_{\ccalH(K)}
.        
\end{multline}
Then, this implies that
\begin{equation}
\rho_{K}
     \left( 
            p(t)
     \right)
             \left(
                      \rho_{K}
                             \left( 
                                   r(t)
                              \right)
                                    f
             \right)  
             =
             \rho_{K}
                   \left( 
                         p(t)
                         r(t)
                   \right)f
                   .
\end{equation}
\end{proof}


\subsection{Proof of Theorem~\ref{thm_alg_rkhs_vs_alg_classic}}
\label{proof_thm_alg_rkhs_vs_alg_classic}
\begin{proof}
From~\eqref{eq_TK_operator} we know that
\begin{equation}\label{eq_proof_thm_alg_rkhs_vs_alg_classic_1}
p\left( 
       \boldsymbol{T}_{K}
 \right)
        f	
=
\sum_{i=1}^{\infty}
          p\left(
                \sigma_{i}
           \right)       
           \vartheta_{i}(v)
           \left\langle 
                 f
                 ,
                 \vartheta_{i}
           \right\rangle_{L_2} 
           .
\end{equation}
Since 
$
\langle 
     f
     ,
     \vartheta_{i}
\rangle_{L_{2}}
=
\sigma_{i}   
\langle 
f
,
\vartheta_{i}
\rangle_{\ccalH(K)}  
$
it follows that
\begin{equation}\label{eq_proof_thm_alg_rkhs_vs_alg_classic_2}
	p\left( 
	      \boldsymbol{T}_{K}
	\right)
          f
=
	\sum_{i=1}^{\infty}
	        p\left(
	              \sigma_{i}
	         \right)     
          \sigma_{i}
	         \vartheta_{i}(v)
	         \left\langle 
                f
                ,
	              \vartheta_{i}
	         \right\rangle_{\ccalH(K)} 
	.
\end{equation}
Now, we recall from the definition of $\rho_{K}$ that
\begin{equation}\label{eq_proof_thm_alg_rkhs_vs_alg_classic_3}
\rho_{K}
	\left( 
	      p\left(
	           t
	       \right)
	\right)
          f
=
    \overline{
	\left\langle 
	      q_{v}\left(K^{\square}\right)
	      ,
	      f
	\right\rangle
    }_{\ccalH(K)}
	,
\end{equation}
and from Theorem~\ref{thm_box_prod_spect} that
\begin{equation}\label{eq_proof_thm_alg_rkhs_vs_alg_classic_4}
q_{v}\left(K^{\square}\right)
=
\sum_{i=1}^{\infty}
     p\left( 
           \sigma_{i}
      \right)
            \sigma_{i}
            \vartheta_{i}(u)
            \overline{\vartheta_{i}(v)}	
            .
\end{equation}
Replacing~\eqref{eq_proof_thm_alg_rkhs_vs_alg_classic_4} into~\eqref{eq_proof_thm_alg_rkhs_vs_alg_classic_3}, and distributing the product (inner) we obtain
\begin{equation}
    \overline{
	\left\langle 
	      q_{v}\left(K^{\square}\right)
	      ,
	      f
	\right\rangle
    }_{\ccalH(K)}
    =
     \overline{
          \sum_{i=1}^{\infty}
                p(\sigma_{i})
                \sigma_{i}
                \overline{\vartheta_{i}(v)}
                \left\langle
                      \vartheta_{i}
                      ,
                      f
                \right\rangle_{\ccalH(K)}
      }
      .
\end{equation}
Using the conjugation operation properties we obtain
\begin{equation}
    \overline{
	\left\langle 
	      q_{v}\left(K^{\square}\right)
	      ,
	      f
	\right\rangle
    }_{\ccalH(K)}
    =
          \sum_{i=1}^{\infty}
                p(\sigma_{i})
                \sigma_{i}
                \overline{
                     \overline{\vartheta_{i}(v)}
                      }
                      ~
                \overline{
                \left\langle
                      \vartheta_{i}
                      ,
                      f
                \right\rangle}_{\ccalH(K)}
      ,
\end{equation}
and since $\overline{\left\langle\vartheta_{i},f\right\rangle}_{\ccalH(K)}=\left\langle f,\vartheta_{i}\right\rangle_{\ccalH(K)}$, it follows that
\begin{equation}
    \overline{
	\left\langle 
	      q_{v}\left(K^{\square}\right)
	      ,
	      f
	\right\rangle
    }_{\ccalH(K)}
    =
          \sum_{i=1}^{\infty}
                p(\sigma_{i})
                \sigma_{i}
                \vartheta_{i}(v)
                \left\langle
                      f
                      ,
                      \vartheta_{i}
                \right\rangle_{\ccalH(K)}
      .
\end{equation}
\end{proof}


\subsection{Proof of Theorem~\ref{thm_TKn_vs_Kboxn}}
\label{proof_thm_TKn_vs_Kboxn}

\begin{proof}
By~\eqref{eq_thm_box_prod_spect_2} we have that
\begin{equation}
     K^{\square n}(u,v)
      =
       \sum_{i=1}^{\infty}
              \sigma_{i}^{n}
              \vartheta_{i}(u)
              \overline{\vartheta_{i}(v)}
              \quad
              \forall n\in\mbN
              .
\end{equation}
Therefore, it trivially follows that $\boldsymbol{T}_{K}^{n}f=\boldsymbol{T}_{K^{\square n}}f$. Now, taking into account that $p(t)\in\mbC[t]$ we can write $p(t)=\sum_{r=0}^{R}h_{r}t^{r}$, which leads to
\begin{equation}
p\left( 
        \boldsymbol{T}_{K}
  \right)
           f
           =
              \sum_{r=0}^{R}
                       h_{r}
                       \boldsymbol{T}_{K}^{r}
                       f
                       .
\end{equation}
Then, it follows that
\begin{equation}
p\left( 
        \boldsymbol{T}_{K}
  \right)
           f
           =
              \sum_{r=0}^{R}
                       h_{r}
                       \boldsymbol{T}_{K^{\square r}}
                       f
           =          
               \sum_{r=0}^{R}
                       \boldsymbol{T}_{h_{r}K^{\square r}}
                       f   
           =
                \boldsymbol{T}_{p\left(K^{\square r}\right)}
                                      f                           
                       .
\end{equation}
\end{proof}


\subsection{Proof of Corollary~\ref{cor_rkhs_uncertainty}}
\label{subsec_cor_rkhs_uncertainty}

\begin{proof}
Let $f\in\ccalH(K)$ be an RKHS-finite signal that is also $B$-bandlimited. Then, there exists a finite $\ccalT\subset\ccalX$ and $\{ a_{t}\in\mbC \}_{t\in\ccalT}$ such that $f = \sum_{t\in\ccalT}a_{t}k_{t}$. Taking into account~\eqref{eq_mercer_K_decomp} we have
\begin{equation}
f(u)  = 
      \sum_{t\in\ccalT}
                a_{t}
                \sum_{i=1}^{\infty}
                \sigma_{i}
                \vartheta_{i}(u) 
                \overline{\vartheta_{i}(t)}
                ,
\end{equation}
which rearranging the sum, leads to
\begin{equation}
f(u)  = 
           \sum_{i=1}^{\infty}
                        \left(
                               \sigma_{i}
                               \sum_{t\in\ccalT}
                                         a_{t}
                                         \overline{\vartheta_{i}(t)}
                        \right)                  
                \vartheta_{i}(u) 
                .
\end{equation}
Then, the $i$-th Fourier coefficient of $f$ is given by
\begin{equation}
\widehat{f}_{i}
         =
           \left\langle
                 f
                 ,
                 \vartheta_{i}
           \right\rangle_{L_{2}}           
           =
              \sigma_{i}
              \sum_{t\in\ccalT}
                        a_{t}
                        \overline{\vartheta_{i}(t)}
            .            
\end{equation}
Since $f$ is also $B$-bandlimited, we must have that
\begin{equation}
              \sigma_{i}
              \sum_{t\in\ccalT}
                        a_{t}
                        \overline{\vartheta_{i}(t)}
                        =
                        0
                        \quad\forall i>B
                        .
\end{equation}
\end{proof}

\end{document}